\documentclass[preprint,12pt,authoryear,times]{elsarticle}

\usepackage[utf8]{inputenc}
\usepackage{amssymb}
\usepackage{lineno}
\usepackage{booktabs}
\usepackage{threeparttable}
\usepackage{caption}
\usepackage{subcaption}
\usepackage{url}
\usepackage{amsmath} 
\usepackage{graphicx}
\usepackage{array}
\usepackage{booktabs}
\usepackage{multirow}
\usepackage{rotating}
\usepackage{makecell}

\journal{}
\usepackage{caption} %
\usepackage{url}
\usepackage{boxedminipage}
\usepackage{textpos}

\journal{Computers, Environment and Urban Systems}

\begin{document}

\author[doa]{Zicheng Fan}

\author[geo]{Chen-Chieh Feng}

\author[doa,dre]{Filip Biljecki\corref{cor1}}

\affiliation[doa]{organization={Department of Architecture, National University of Singapore}, country={Singapore}%
            }

\affiliation[geo]{organization={Department of Geography, National University of Singapore}, country={Singapore}%
            }
            
\affiliation[dre]{organization={Department of Real Estate, National University of Singapore}, country={Singapore}%
            }                  
            
\cortext[cor1]{Corresponding author}

\begin{frontmatter}

\title{Coverage and bias of street view imagery in mapping the urban environment}

\begin{abstract} 
\begin{textblock*}{\textwidth}(0cm,-11cm)
\begin{center}
\begin{footnotesize}
\begin{boxedminipage}{1\textwidth}
This is the Accepted Manuscript version of an article published by Elsevier in the journal \emph{Computers, Environment and Urban Systems} in 2025, which is available at:\\ \url{https://doi.org/10.1016/j.compenvurbsys.2025.102253}\\ Cite as:
Fan Z, Feng CC, Biljecki F (2025): Coverage and bias of street view imagery in mapping the urban environment. \textit{Computers, Environment and Urban Systems}, 117: 102253.
\end{boxedminipage}
\end{footnotesize}
\end{center}
\end{textblock*}

\begin{textblock*}{1.5\textwidth}(-0.8cm,16.3cm)
{\tiny{\copyright{ }2025, Elsevier. Licensed under the Creative Commons Attribution-NonCommercial-NoDerivatives 4.0 International (\url{http://creativecommons.org/licenses/by-nc-nd/4.0/})}}
\end{textblock*}
Street View Imagery (SVI) has emerged as a valuable data form in urban studies, enabling new ways to map and sense urban environments. However, fundamental concerns regarding the representativeness, quality, and reliability of SVI remain underexplored, e.g.\ to what extent can cities be captured by such data and do data gaps result in bias. This research, positioned at the intersection of spatial data quality and urban analytics, addresses these concerns by proposing a novel and effective method to estimate SVI's element-level coverage in the urban environment. The method integrates the positional relationships between SVI and target elements, as well as the impact of physical obstructions. Expanding the domain of data quality to SVI, we introduce an indicator system that evaluates the extent of coverage, focusing on the completeness and frequency dimensions. Taking London as a case study, three experiments are conducted to identify potential biases in SVI's ability to cover and represent urban environmental elements, using building facades as an example. 
It is found that despite their high availability along urban road networks, Google Street View covers only 62.4 \% of buildings in the case study area. The average facade coverage per building is 12.4 \%. SVI tends to over-represent non-residential buildings, thus possibly resulting in biased analyses, and its coverage of environmental elements is position-dependent. The research also highlights the 
variability of SVI coverage under different data acquisition practices and proposes an optimal sampling interval range of 50--60 m for SVI collection. %
The findings suggest that while SVI offers valuable insights, it is no panacea --- its application in urban research requires careful consideration of data coverage and element-level representativeness to ensure reliable results.

\end{abstract}


\end{frontmatter}

\section{Introduction}
\label{sec:sample1}

Street View Imagery (SVI) has gained a significant role in urban studies and in spatial data infrastructure as a new means to map and sense urban environments~\citep{biljecki_street_2021,kang_review_2020,ibrahim_understanding_2020,Zhang2024}.
Research efforts have been predominantly focused on the development of use cases, while fundamental concerns of data quality and reliability of this emerging form of data have not been given sufficient attention in international scientific literature. 
The lack of understanding of questions such as reach and coverage of SVI data may %
have adverse effects on use cases and downstream analyses.
For example, SVI has been used intensively for mapping street greenery~\citep{zhu_utilizing_2023,liu_establishing_2023} and buildings~\citep{zhong_city-scale_2021,Ramalingam2025}, assessing walking environment~\citep{liu_detecting_2023,he_using_2023} and microclimate~\citep{fujiwara_microclimate_2024}, and understanding human perception at the urban scale~\citep{wu_using_2023,wang_measuring_2022,ramirez_measuring_2021,verma_predicting_2020}, but not much is known about the representativeness and suitability of the data for the corresponding road and sidewalk scenarios, or for the investigated neighborhoods and local zones, e.g.\ it is not known what is the reach of data and to what extent can we sense an urban aspect using SVI, and whether the (incomplete) coverage is representative or biased.

This challenge is exacerbated by the gap that traditional spatial data quality metrics, such as accuracy and resolution, are primarily designed for remote sensing imagery or geometric data, which do not fully apply to SVI~\citep{hou_comprehensive_2022}. While some studies employ completeness to evaluate the integrity of SVI in geospatial coverage~\citep{kim_examination_2023,hou_comprehensive_2022, juhasz_user_2016,quinn_every_2019}, they predominantly focus on the spatial and temporal availability of imagery, limiting their analysis to their coordinates and timestamps. However, there are other unique metadata and properties that set SVI apart from traditional urban data forms, leading to potential variability in its application. For example, the impact of SVI camera parameters (e.g., heading, Field of View (FOV), image formats), and collection intervals are frequently neglected in common SVI-based research practices \citep{kim_decoding_2021,biljecki_sensitivity_2023}. These metadata shape the potential of SVI to map specific street elements horizontally in urban environments, which are crucial for SVI as a proxy for human-centered sensing and perception. To address some of these limitations, \citet{hou_comprehensive_2022} propose a comprehensive framework to evaluate SVI data quality, primarily focusing on image quality, metadata availability and accuracy, and spatial and temporal aspects. The framework promotes SVI metadata and improves the standardization in SVI collection and utilization, especially for volunteered street view imagery (VSVI), which are subject to heterogeneous acquisition practices~\citep{hou_comprehensive_2022,Danish2025,Helbich2024}. However, the framework is tightly constrained to quality considerations, with limited exploration of how metadata practically impacts SVI's capability to represent environmental information.

Even with high quality of imagery, homogeneous acquisition protocols employed by commercial providers, extensive availability, and proper metadata control, SVI does not necessarily guarantee reliable representation of the urban environment. One major limitation is that SVI is largely constrained to roads. It is reported that SVI has diminishing reach from public streets to the interior roads of blocks or neighborhoods~\citep{biljecki_street_2021,kang_review_2020}. Moreover, the complexity of real-world objects and their layouts, as captured by SVI, can introduce noise that impacts the reliability of this data in covering specific environmental elements. For example, similarly as how vegetation and clouds can obstruct the observation of ground-level objects in remote sensing imagery~\citep{hosseini_mapping_2023}, elements such as trees and vehicles may obstruct street-level mapping of urban elements such as building facades~\citep{novack_towards_2020}. Their interplay is also important --- buildings, while often the main focus of use cases, can also act as unwanted obstacles to other objects~\citep{yan_estimation_2022,raghu_towards_2023}. In densely built environments with congested layouts, such mutual obstructions are further amplified, which hinders SVI from providing complete and extensive coverage of environmental elements. Efforts to address these challenges include the use of generative models to inpaint obstructed SVI images, such as removing trees, street furniture from building facades \citep{yu_inpaint_2023,hu_saliency-guided_2023}. However, the systematic effects of these obstructions on urban sensing using SVI remain untouched. As a result, the complexity of urban environments should be considered another critical concern.

In spite of the importance, there is a notable absence of effective methodology and set of metrics to assess the SVI's representativeness of environmental information, not just relying on a general coverage analysis inferred from SVI locations but also by diving into \textit{\textbf{element-level coverage}}. The assessment should examine the discrepancies and biases between SVI covered element information to its real-world distribution, and address the uncertainty introduced by different SVI metadata, their practical usage, and the complexity of urban environment.  
Some existing studies only explore the stability and sensitivity of SVI in proxying environmental elements (e.g. building, sky, and greenery), and under varying metadata settings such as different image formats and projection methods \citep{biljecki_sensitivity_2023}, different image collection intervals \citep{kim_examination_2023}, from pedestrian and vehicle perspectives and different directions~\citep{ki_bridging_2023}. However, whether stable or not, the degree to which SVI-based environmental measurement correspond to actual environmental elements is still undetermined.
This lack of knowledge suggests a need for further and more surgical research that departs from the general SVI data quality studies or sensitivity research, to understand the extent and limitations of SVI in providing a comprehensive view of urban environments.

With the above elaborated gaps and research ideas, the main research questions we seek to answer in this paper are:
\begin{quotation}

    \noindent Q1. How can we quantitatively estimate and describe the element-level coverage of SVI on urban environment? \\
    
    \noindent Q2. Are there typical biases or discrepancies in SVI's representation of the urban environment when analyzed through the element-level perspectives?

\end{quotation}

To address the questions, the research proposes a novel workflow to estimate SVI's coverage on elements in urban environment. The workflow integrates both the positional relationships between SVI and the target element, and the obstructions from environmental objects and settings into consideration, applying isovist analysis and semantic segmentation methods. Moreover, an accompanying indicator system is developed to evaluate and describe the coverage extent. Key considerations include the degree to which total street elements in a city can be captured in SVI, and whether certain instances in the element are repeatedly covered while others are frequently left out of sight.
Taking the central area in the Greater London as the case study area, the research is further structured around three experiments to identify potential bias of SVI in covering and representing environment information. Urban building facade is selected as the example element for these experiments. In Experiment 1, we examine the distribution characteristics of building information captured by SVI and compare it with the initial distribution based on building footprint data. Experiment 2 focuses on SVI coverage at the aggregated level, comparing the proposed element-level coverage estimates with traditional coverage estimates that rely on spatial distribution. In Experiment 3, we explore the impact of different SVI collection intervals, an important aspect of SVI metadata, on the variability of element-level SVI coverage. Through the experiments above, we justify our element-level SVI coverage estimation workflow and metrics, and offer useful suggestions and reference in improving the reliability of further SVI-based urban research.

\section{Background and related work}

\subsection{Application and concerns of SVI in mapping urban environment elements}

Thanks to the rapid advancements in deep learning, particularly in visual tasks such as image classification, semantic segmentation, and object detection, Street View Imagery (SVI) data has been widely applied in urban science. By combining SVI with deep learning, researchers have been able to map and classify urban elements such as buildings, roads, and greenery on a large scale, generating new geospatial data or enhancing existing datasets \citep{biljecki_street_2021,seiferling_green_2017}. SVI also serves as a visual proxy for investigating socio-economic attributes and human perceptions of specific neighborhoods or urban spaces \citep{he_urban_2021,fan_urban_2023, kang_review_2020}. Table \ref{table:urban_features} summarizes recent studies that use SVI to map and sense typical environmental elements.

The considerable quantity and the extensive distribution of SVI, coupled with scalable machine learning models, are the prominent reasons SVI has been a popular urban data source for city-scale analyses \citep{biljecki_street_2021,kang_review_2020}. Nevertheless, existing research falls short in articulating the specific extent of SVI's representativeness. Specifically, in terms of mapping urban environment elements such as building and greenery, it is questionable to which degree SVI can reach all the corresponding elements compared to their nature of existence. For perception research relying SVI as visual proxy, similar concerns raise about whether the SVI captured information, shares a similar distribution with the information represented in total investigated scenarios, or in the selected regions or neighborhoods. Acknowledging the potential limitations, models trained on SVI for mapping environmental elements or evaluating spatial attributes are often presented as an initial step or baseline for a broader research goal \citep{raghu_towards_2023, yan_estimation_2022, ramalingam_automatizing_2023}. Fully achieving the goal requires addressing the uncertainty in SVI's ability to systematically represent the urban environment.

\subsection{Sources of uncertainty in SVI application}
To better understand the limitations of SVI, the section identifies three typical sources of uncertainty reported in previous research, namely SVI data availability and quality, common practices in SVI utilization, and complexity in real-world environment.

\subsubsection{SVI data quality: availability, image quality, and others}
SVI data quality problems can be regarded as an inherent source of uncertainties for SVI in representing urban environment. Among them, the data availability have received more attention, and there are uneven availability distribution of SVI images and services across various geographic scales. 
Notably, a significant number of cities worldwide still lack SVI services. Cities in Europe and North America enjoy broader SVI coverage, whereas economically underdeveloped regions in Latin America and Africa experience sparse and limited distribution \citep{bendixen_putting_2023,hou_comprehensive_2022, quinn_every_2019}. Moreover, there is spatial heterogeneity of SVI collection within cities where SVI services are available. For example, streets in areas characterized by high traffic volumes, dense populations, or wealthier demographics are more likely to be imaged \citep{fry_assessing_2020}. Conversely, smaller towns and rural areas are often overlooked \citep{szczepanska_evaluation_2020}. Additionally, image availability in informal urban sectors can be compromised by the absence of accessible roads \citep{chen_multi-modal_2022,he_urban_2021}, which results in the structured missingness problem \citep{mitra_learning_2023}.
In their review papers, \citet{kang_review_2020} and \citet{biljecki_street_2021} point out that SVI is predominantly collected along streets, making it challenging to analyze variations within neighborhood built environments, potentially leading to structural issues in the completeness of information.

The distribution of SVI availability also varies among different service providers. 
Google Street View often employs an \textit{all-or-nothing} approach to collecting SVI \citep{quinn_every_2019}, and achieves a more complete coverage on road network for cities where the service is available. 
In contrast,crowdsourcing platforms such as Mapillary, which rely on user contributions, lag behind commercial services in terms of the number of globally available cities and the completeness of road coverage \citep{biljecki_street_2021,wang_investigating_2024}. Nevertheless, \citet{zheng_does_2024} found that as cumulative image contributions along streets increase, the spatial density and viewing angle coverage of VSVI improve significantly. Additionally, the temporal continuity of SVI availability in same or nearby geo-locations is a concern for both commercial and crowdsourced sources \citep{kim_examination_2023,hou_comprehensive_2022}. 
There are series of method for measuring spatial and temporal availability of SVI data. For Google Street View, a typical method for assessing availability is to extract sampling points randomly or in a standardized way based on the road network in specific urban areas, calculating the proportion of these sampling points that have valid SVI in the vicinity \citep{smith_google_2021,fry_assessing_2020,kim_examination_2023}. For crowdsourced data such as Mapillary or KartaView, besides the method above, the availability of SVI can also be measured by the proportion of road lengths covered by continuous SVI sequences to the total length of the road network \citep{mahabir_crowdsourcing_2020,hou_comprehensive_2022}. Spatial coordinates and timestamps in SVI metadata play a crucial role in these assessment methods. 
As this paper will demonstrate, such methods may be useful but simplistic, so they may not paint a complete picture of the coverage of SVI, especially in the context of use cases.
For example, it is difficult to tell from such metrics what is the percentage of buildings that can be mapped from SVI.

Beyond general data availability problems, SVI also suffers from image quality problems, such as blurriness, and variable lighting and weather conditions \citep{rui_quantifying_2023,zou_mapping_2022,yuan_using_2023}. Low-quality and defective images may hinder the performance of computer vision models \citep{vo_search_2023}. The missing or inaccuracy of SVI metadata, such as GPS coordinates, timestamps, and exterior orientation parameters can also limit the usability of SVI in reflecting environmental information \citep{lumnitz_mapping_2021,liang_automatic_2017}. Given the situation, \citet{hou_comprehensive_2022} first proposed a comprehensive framework to assess SVI quality problems beyond availability. The quality issues are conceptualized into 48 elements across 7 categories, namely image quality, metadata availability and accuracy, spatial quality, temporal quality, logical consistency, redundancy, and privacy. 
The relevant evaluation system and methods have been used to create an open global street view dataset, with a focus on enhancing metadata in existing crowdsourced street view data sources \citep{hou_global_2024}. %

\subsubsection{Common practices in SVI utilization}
Beyond data availability and quality, the way researchers utilize SVI data introduces another layer of uncertainty. %
As a common practice, many studies retrieve SVI by sampling points at regular intervals along road networks, selecting the nearest images for analysis. This approach primarily aims to mitigate potential spatial unevenness in SVI distribution. 

On the one hand, the road networks used for sampling, often based on OpenStreetMap (OSM), may have limitations in terms of timeliness and completeness \citep{sanchez_accessing_2024}. In underdeveloped areas or regions with policy restrictions, the coverage may be even less comprehensive. This issue can be amplified by SVI's focus on main streets, often neglecting pedestrian paths or internal neighborhood spaces \citep{kang_review_2020, biljecki_street_2021}. Given the situation, research focused on mapping trees and plants usually restricts the mapped targets to those located along the street \citep{lumnitz_mapping_2021, liang_characterizing_2024}. However, for studies primarily concerned with mapping buildings, the spatial boundaries of the targeted objects are often vaguely defined \citep{zhou_evaluating_2023,aravena_pelizari_automated_2021}. 

On the other hand, there is a lack of sufficient evidence to determine the optimal interval for SVI sampling. Smaller intervals may introduce redundancy, which is advantageous for mapping environmental elements as it ensures the capture of useful information and prevents data gaps \citep{liang_characterizing_2024}. However, in studies focused on spatial perception at the neighborhood or regional level, overly dense intervals could result in the repeated capture and overemphasizing of certain environmental element in the overall captured information, leading to a biased representation. \citet{kim_decoding_2021} systematically examined how different sampling intervals affect the SVI-based measurement of various street view elements, finding significant fluctuations across intervals. However, their focus was more on the stability of these measurements rather than on how accurately SVI reflects the real environment.

Beyond the image sampling methods and intervals, researchers also consider the impact from other parameter selections and practices on SVI utilization. Specifically, these studies have compared different SVI orientations \citep{kim_decoding_2021}, different collection positions (lanes and sidewalks) \citep{ki_bridging_2023,ito_translating_2024}, and different image sources and forms (crowdsourced and commercial; perspective and panoramic) \citep{biljecki_sensitivity_2023}, to see if they affect how SVI summarizes and reflects the same urban environment elements, such as buildings, greenery, and sky.
It is reported that though reliability of a single, crowdsourced imagery is comparable to commercial panoramas \citep{biljecki_sensitivity_2023}, there are significant measurement errors for sidewalks, greenery, and roads between pedestrian and vehicle views \citep{ki_bridging_2023}. 
Additionally, \citet{liu_clarity_2024} discussed the issues such as lack of clear technical definitions in street attributes extracted from SVI, the lag in CV model performance, and the absence of benchmarks. Although not the focus of this paper, these issues contribute another aspect of uncertainty in the common practices of SVI utilization.

\subsubsection{Complexity of real-world environment}
Different sampling strategies and parameter selections for SVI primarily test the stability of SVI in mapping diverse urban environments. Beyond stability, however, the extent to which SVI can effectively cover and represent complex environments remains under-researched. The complexity of urban environments can be further explained by the obstructions of other environmental instances, and the density and layout of the overall built environment and their heterogeneity. \citet{biljecki_street_2021} and \citet{novack_towards_2020} note that objects frequently analyzed in SVI-based studies, such as buildings and trees, are often blocked by other street-level elements in the imagery. Specifically, obstructions are one of the main issues when using SVI to measure tree size \citep{liang_characterizing_2024}, estimate building height \citep{yan_estimation_2022} and classify building materials \citep{raghu_towards_2023}. The presence of obstructions also amplifies the differences in environmental measurements based on pedestrian versus vehicle perspectives \citep{ki_bridging_2023}.

Additionally, when mapping buildings or trees in urban environment, obstructions can also be caused by the targeted elements themselves \citep{yan_estimation_2022,raghu_towards_2023}, which is related to the density and layout distributions of element instances in the surrounding environment. Obstruction is fundamentally a problem of the relative position between the camera and the target element instances. While SVI can be collected using relatively standardized procedures, the distance and angle between environmental elements and the camera lens can vary significantly \citep{zou_mapping_2022,lumnitz_mapping_2021,huang_no_2025}. 
In studies focused on building assessments, some retrieved SVI may only show the sides or partial views of buildings due to the variation of horizontal angles, which might not provide sufficient façade features for classification or evaluation tasks \citep{zou_mapping_2022}. Furthermore, suburban neighborhoods with sparse, wide roads may differ significantly in openness from densely packed commercial areas in city centers. This will affect not only the quantity and distribution of SVI collected, but also the completeness and frequency with which SVI captures specific environmental elements and instances.

In summary, the complexity of the urban environment introduces additional uncertainty in SVI coverage of environmental elements. However, compared to data quality problems or impact from SVI usage practices discussed in previous sections, it has not been given enough attention. There is a notable research gap concerning the extent to which SVI accurately covers the spatial instances in the environment, corresponding to the visual elements it aims to represent in the image space. Limited and relevant examples are only about the greenness visibility \citep{yan_evaluating_2023,labib_modelling_2021}, where Green View Index (from SVI) and Viewshed Greenness Visibility Index (from GIS simulation) are compared. The focuses are about the similarity and discrepancies between the two indicators, rather than element coverage potential of SVI.
For this reason, new perspective and method are explored to estimate the SVI coverage on urban environmental elements in this study, quantitatively incorporating the impact from environmental complexity. 
The details are depicted in the following sections.

\section{Methodology}

\begin{figure}[htbp]
    \centering
    \begin{subfigure}{1\textwidth}
        \centering
        \includegraphics[width=\textwidth]{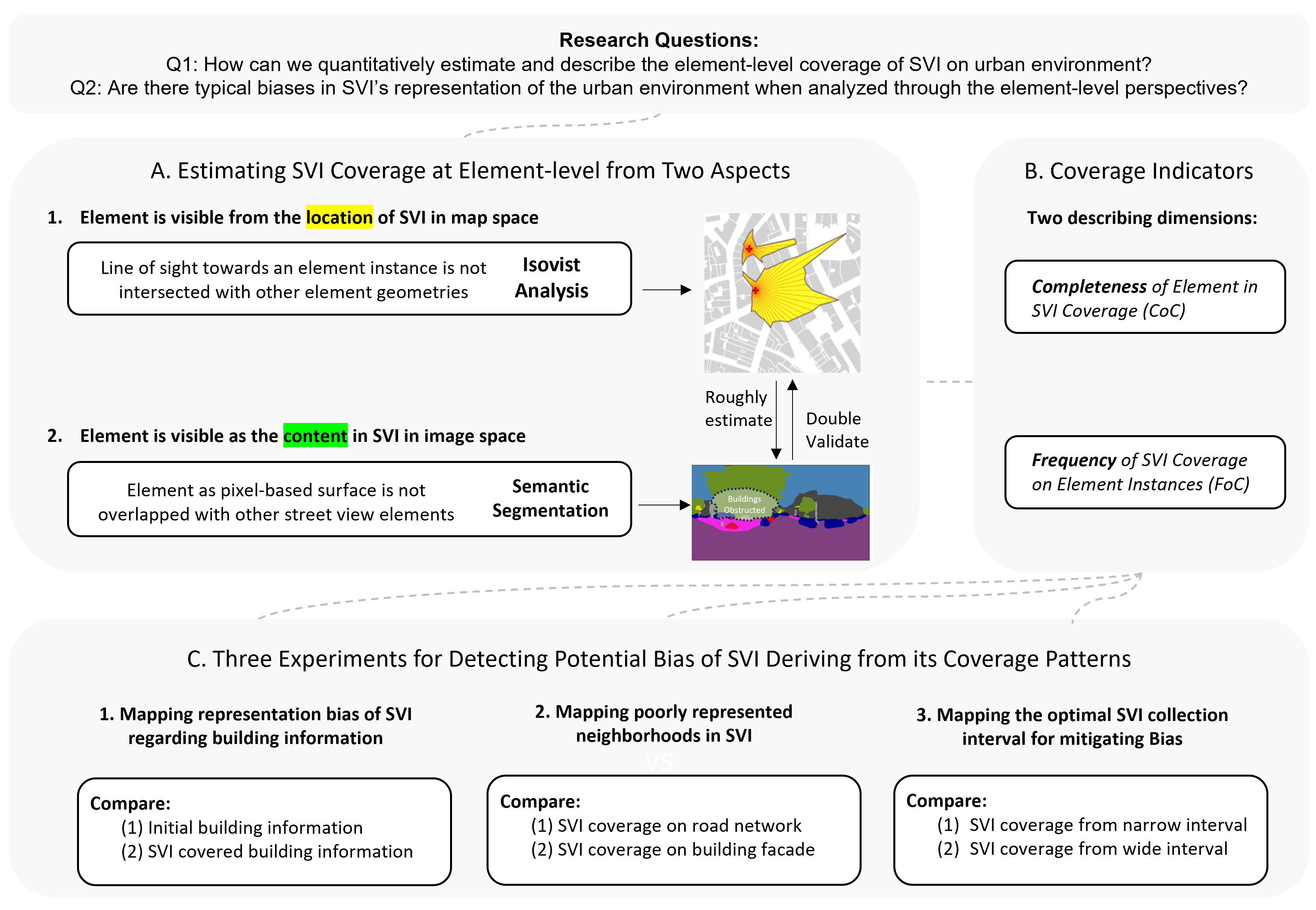}
    \end{subfigure}
    \caption{Research framework. The isovist analysis example is generated based on `t4gpd' Python library.}
    \label{fig:research framework}
\end{figure}

A research framework of the study is illustrated in Figure \ref{fig:research framework}. We propose a novel method to estimate element-level coverage of SVI,
that integrates isovist analysis method developed by \cite{benedikt_take_1979} and computer vision technologies. On this basis, comprehensive SVI coverage indicators are designed and calculated to describe the SVI coverage extents in different dimensions and in multiple geographical scales. Utilizing the SVI coverage indicators, we design three experiments to identify the potential bias of SVI in representing built environment information in horizontal dimension.

\subsection{Element-level SVI coverage estimation}

\subsubsection{Concept}

As SVI can be regarded as the projection of 3D urban environment onto 2D image space at specific locations, the coverage of SVI on environment elements can be naturally examined both from two perspectives: from the visibility of environmental elements in image space, and from the relationship between SVI locations and element locations in the geometric space. The dual relationship forms the foundation of the proposed SVI coverage estimation method, which is sufficiently adaptable to various elements of the built environment. 

In this study, urban building facades are chosen as a representative element for coverage estimation and bias assessment. This choice is motivated by the fact that buildings typically account for a significant portion of the visual information captured in SVI and also serve as the primary containers of urban functions and activities. Mapping buildings as static elements in urban environment is also more common and reliable in current research practices compared to mapping dynamic elements, such as pedestrians. %
In addition, by exploring how other environmental elements, such as trees or vehicles, obstruct SVI’s coverage on building facades, this investigation sheds light on how the complexity of urban environment shape the utility and limitations of SVI. 
A two-step workflow for estimating SVI coverage on building facades is introduced in the following sections.

\subsubsection{Step 1 -- coverage estimation based on geometric analysis}
The SVI coverage on building facades can be first defined in 2D space, as the intersection of the visual field of a potential observer at SVI location, with respect to the surrounding building instances. The definition is based on the nature of SVI as the collection of visible street elements at specific geographical locations. Practically, the SVI coverage can be computed quantitatively based on the isovist analysis method, as the proportion of building facades directly visible from SVI locations and within a given distance. Buildings serve as both the observed objects and the visual obstacles in the analysis, and how frequently and how completely the building facades can be visible from SVI locations, represent the extent building covered by SVIs.

\begin{figure}[htbp]
    \centering
    \begin{subfigure}{1\textwidth}
        \centering
        \includegraphics[scale=0.4]{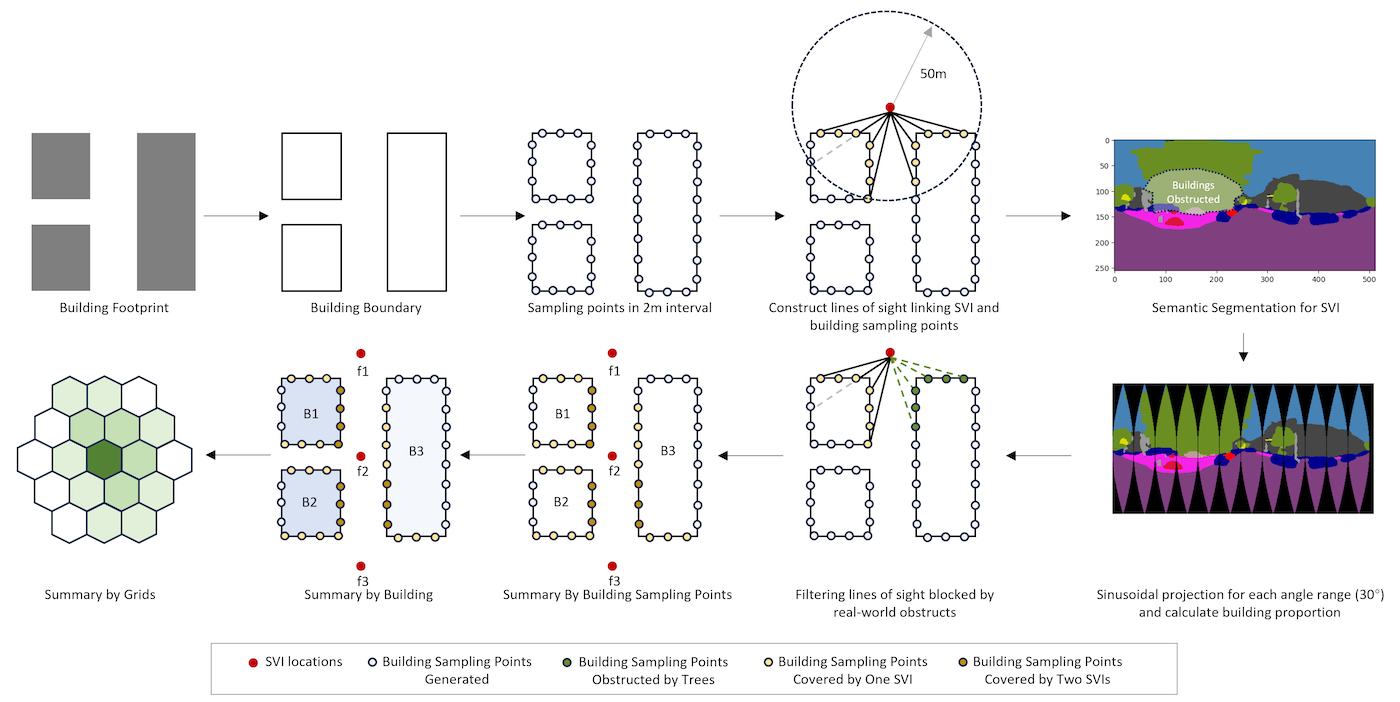}
    \end{subfigure}
    \caption{A simplified workflow integrating isovist analysis and computer vision technology for estimating the SVI coverage on building facades.}
    \label{fig:isovist_workflow}
\end{figure}

A Python script is designed to carry out simplified isovist analysis for SVI location points. As shown in the workflow illustrated in Figure \ref{fig:isovist_workflow}, sampling points in 2 m interval are extracted from the boundaries of building footprints, as the unit representation of building facades which are potentially visible. A threshold of 50 m is set for isovist analysis, as the proximity of the maximum distance where human can achieve an efficient observation in complex urban environment. For each SVI location, lines of sight are first constructed towards all the building samplings points within the distance threshold. The spatial join method in `GeoPandas' Python library is then applied to filter lines of sight which are not intersected with the surrounding building footprints. Each line of sight filtered counts as once a building sampling point can be seen via a specific SVI location. While there are existing tools to carry out 2D isovist analysis \citep{leduc_t4gpd_2024}, such as `t4gpd',\footnote{\url{https://github.com/thomas-leduc/t4gpd/}} they are not fully applicable in our work due to being computationally intensive for this kind of analysis. Thus, we develop our own implementation.

\subsubsection{Step 2 -- coverage validation based on image content}
To validate whether the lines of sight are blocked by non-building elements in the real street environment, we calculate the absolute angles for lines of sight compared to the true north, and relocate them in the image space of SVI, with the SVI metadata of location and heading. For each SVI, semantic segmentation is conducted via the Python library `ZenSVI',\footnote{\url{https://github.com/koito19960406/ZenSVI/}} to detect different street view elements, such as building, road, sky, vehicle and greenery \citep{ito_zensvi_2024}. Then the SVI as a panoramic image is horizontally divided into 12 pieces with equal angle range. Sinusoidal projection is applied to each angle range to restore the distortion of street view elements at polar points and suit for an eye-level view. The proportion of building elements with respect to the non-sky and non-ground elements in the angle range are calculated, as shown in Formula \ref{formula:proportion}. For angle ranges with building proportion below a certain threshold, the lines of sight within the ranges can be removed, as they are highly likely to be blocked by other street obstacles. By summarizing the remaining lines of sight by building footprints and by local geospatial units, we gain a cumulative distribution of SVI coverage potential. More comprehensive indicators can be calculated to describe the SVI coverage on building facades in different dimensions.

\subsection{SVI coverage indicators}

We propose a novel indicator system to describe the extent of element-level SVI coverage 
in two dimensions, namely the completeness and frequency. %
Taking SVI coverage on a single building as an example, the completeness indicator measures how thoroughly the open facades of a building can be viewed from surrounding panoramic SVIs. Specifically, the \textit{Completeness of SVI Coverage for Individual Building} (\textit{CoC-B}) is given by the Formula \ref{eq:exposure}, as the ratio of the visible building sampling points with respect to all the sampling points that are available from public facades for a single building. \textit{CoC-B} reflects the representativeness of building information captured by SVI compared to what building conveys in public. In terms of frequency, the indicator represents the total number of lines of sight that reach one building and from the surrounding panoramic SVIs. Lines of sight between the same SVI and building pair, but through different building sampling points are counted as independent occurrences. Since larger buildings with longer facades naturally have a higher probability of being viewed, these occurrences are weighted by the building's perimeter. The \textit{Frequency of SVI Coverage on A Single Building} (\textit{FoC-B}) is given by the Formula \ref{eq:frequency}. A higher \textit{FoC-B} indicates there is a higher probability a building can be viewed in urban environment.

\begin{equation}
\textit{CoC-B} = \frac{U_{\text{seen}}}{U_{\text{avail}}}
\label{eq:exposure}
\end{equation}
where
\begin{itemize}
    \item $U_{\text{seen}}$ denotes the actual number of unique sampling points visible from the surrounding panoramic SVIs.
    \item $U_{\text{avail}}$ represents the total number of unique sampling points available around the building, providing a measure of the potential for SVI coverage.

\end{itemize}

\begin{equation}
\textit{FoC-B} = \frac{V}{P}
\label{eq:frequency}
\end{equation}
where
\begin{itemize}
    \item $V$ represents the occurrences sampling points from one building being visible by panoramic SVIs in the surrounding.
    \item $P$ denotes the building perimeter, serving as a measure for normalizing visibility by the building's size.
\end{itemize}

Beyond the building-level indicators, we also design indicators to describe the SVI coverage extent at the aggregated level, similarly in the dimensions of frequency and completeness. Specifically, the \textit{Completeness of SVI Coverage on Buildings in Local Area (CoC-A)}, is designed to describe the proportion of buildings with at least one line of sight reached in local areas, such as neighborhoods or census units. The indicator is given by the Formula \ref{eq:completeness_area}. The indicator can be applied to detecting areas with insufficient SVI coverage from building perspective, or conversely, evaluating the privacy risk of neighborhoods when exposed to SVI. The \textit{Frequency of SVI Coverage on Buildings in Local Area (FoC-A)} denotes the proportion of SVI coverage occurrence on a certain building type, relative to the total SVI coverage occurrence across all the building types in the local area. Given by the Formula \ref{eq:frequency_area}, this indicator plays a significant role in measuring the impact of a building type on the overall character and visual perception of an area. A higher concentration of SVI coverage of specific building types within an area suggests that renovations and improvements to buildings of this type could potentially have greater visual and social impacts. Table \ref{tab:svi_coverage} in the appendix provides a summary for all the four SVI coverage indicators.

\begin{equation}
\textit{CoC-A} = \frac{N_{\text{seen}}}{N_{\text{total}}}
\label{eq:completeness_area}
\end{equation}
where
\begin{itemize}
    \item $N_{\text{seen}}$ denotes the number of buildings with SVI coverage in the local area.
    \item $N_{\text{total}}$ represents the total number of buildings in the local area.
\end{itemize}

\begin{equation}
\textit{FoC-A} = \frac{\sum_{i} V_{i,\text{type}}}{\sum_{j} V_{j,\text{total}}}
\label{eq:frequency_area}
\end{equation}
where
\begin{itemize}
    \item $V_{i,\text{type}}$ represents the SVI coverage occurrence for the $i^{th}$ building in a specific building type in the local area.
    \item $V_{j,\text{total}}$ represents the SVI coverage occurrence for the $j^{th}$ building across all building types in the local area.
\end{itemize}

\section{Case study}

A case study is conducted in Greater London, UK, to implement the proposed workflow for SVI coverage estimation. Following the results of the SVI coverage estimation, the study undertakes three experiments to understand the potential bias and uncertainty related to SVI coverage and potential data gaps. Further solutions and suggestions are provided based on the experiments to help support the robust application of SVI in urban research. 

\subsection{Data collection}

The largest and most popular commercial street view service -- Google Street View (GSV) is selected as the source of SVI data in this case study. We achieve a thorough search for all the latest SVI locations in the Greater London administrative area, via the official Google Maps API and the python library `streetlevel'.\footnote{\url{https://github.com/sk-zk/streetlevel/}} `Streetlevel' provides a feasible method to fetch all the available SVI locations by map tiles. The method can succeed the traditional SVI collection method relying on road sampling points, which may result in an incomplete SVI searching. %
Totally, 2,590,604 SVI location points and their heading directions are collected in the Greater London administrative area.  We randomly sampled 1 \% of these location points and analyzed their nearest distances to neighboring points. It was revealed that over 71.3 \% of the sampled points had at least one neighboring point within a 10-m buffer.
In addition to SVI data, the research adapts the building footprint and road network data from OSM as the representation of urban environment elements. Land use data from the Colouring Cities Research Programme (CCRP) \citep{hudson2024ccrp} is applied to supplement building data from OSM, providing representations of building functional types, since such information is not always available in OSM~\citep{2023_bae_osm_qa}. Detailed classification information is available in \ref{sec:sample:appendix_D}. %

\subsection{Data management}

\begin{figure}[htbp]
    \centering
    \begin{subfigure}{1\textwidth}
        \centering
        \includegraphics[scale=0.4]{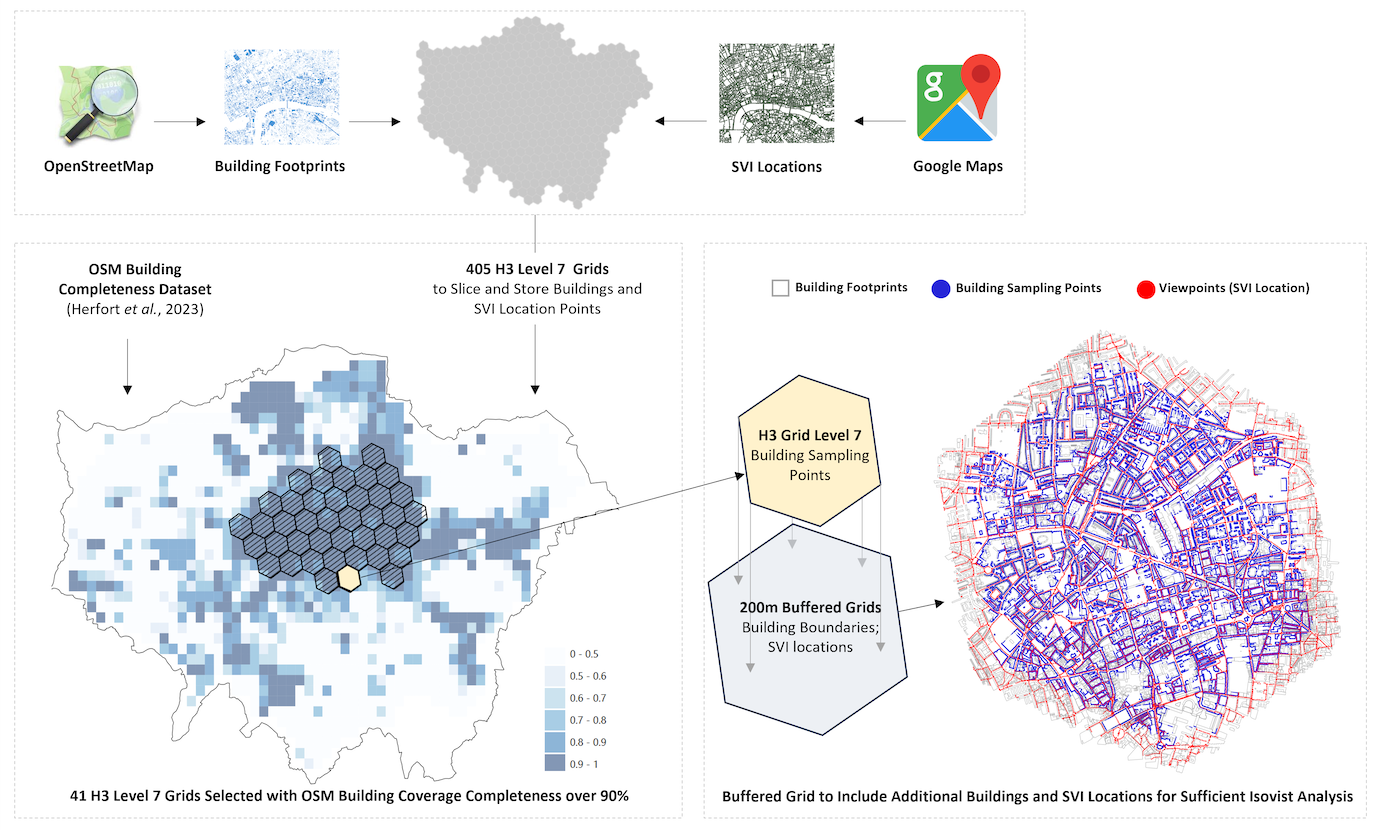}
    \end{subfigure}
    \caption{The data collection and pre-processing workflow in the case study. Source of the base map data: OpenStreetMap, Greater London Authority.}
    \label{fig:data_indexing}
\end{figure}

Considering that the isovist analysis is a computation intensive analytics method, and the datasets collected above contain a large quantity of geometry shapes, which can bring challenges to data loading and processing, the research employs Uber’s H3 discrete global grid system (DGGS)\footnote{\url{ https://h3geo.org/}} to manage the datasets in an efficient manner. The H3 DGGS adopts a hierarchical representation to divide Earth’s surface into grids at sixteen different resolutions. As illustrated in Figure \ref{fig:data_indexing}, the H3 level-7 grids, which have an average hexagon area of 5.16 km$^{2}$ and an average edge length of 1.40 km, are applied to split SVI locations, building footprints and sampling points into smaller groups for isovist analysis. Each building sampling point is assigned a unique H3 id and building id for further aggregation and statistics analysis. To avoid the incomplete analysis for building samplings points at the edge of grid, SVI location points and building footprints are indexed in an extended 200 m buffer area for each H3 level-7 grid, to ensure building sampling points are surrounded sufficiently with potential observers and obstacles in the isovist analysis.
Given that OSM building footprint data may not always be complete~\citep{herfort_spatio-temporal_2023}, we select 41 H3 level-7 grids with building footprint completeness over 90 \% as case areas in study. The building completeness information is referenced from the open dataset released from the work by \citet{herfort_spatio-temporal_2023}.

\subsection{Experiments}

\subsubsection {Identifying potential bias in the SVI covered building information}

For the Experiment 1, the study aims to detect the potential bias of SVI in representing building information. To begin with, we hope to know whether SVI coverage achieves an even distribution across buildings of different function types and sizes, or it is prone to highlight or ignore specific buildings in the built environment. The experiment is conducted by comparing interior distribution of the building-level SVI coverage completeness indicator, \textit{CoC-B}, across buildings grouped by building types and sizes. 

Furthermore, we hope to learn whether the SVI captured building information is representative of the building in reality.
For building function types such as residential, retail, transport etc., we calculate the area-level frequency indicator, \textit{FoC-A}, and their proportion by count in OSM in each H3 level-9 grid. By exploring the linear association between the two indicators, especially interpreting the correlation scores and the regression coefficients, we investigate whether specific building function types are prone to be over-represented or under-represented in SVI, i.e.\ whether using SVI to map the built environment is biased.

\subsubsection {Mapping and explaining poorly represented neighborhoods in SVI}

In the Experiment 2, we aim to identify neighborhoods that are not adequately represented in SVI. This investigation is based on the hypothesis that even if SVI provides sufficient coverage in terms of spatial distribution within a neighborhood, it may still lack adequate coverage on building facades or on other environmental elements. Consequently, this insufficiency might hinder a comprehensive representation of the built environment within the neighborhood. Examining this hypothesis could help uncover potential biases in numerous neighborhood-focused studies based on SVI data.

The experiment is conducted by comparing the spatial distribution of traditionally adopted SVI coverage indicator with the spatial distribution of new coverage indicators proposed in this study. 
Specifically, for each H3 level-9 grids in the case study area, we aggregate the mean values of \textit{CoC-B} indicators and calculate the \textit{CoC-A} indicators, as proxies of SVI's capability in covering building facades in local areas. Concurrently, referring to previous work by \citet{juhasz_user_2016} and \citet{hou_comprehensive_2022}, we compute the completeness of SVI coverage on road networks in the H3 level-9 grids as benchmarks. %
The calculation detail is described in \ref{sec:sample:appendix_E}.
Utilizing Getis-Ord Gi* analysis, a spatial auto-correlation analysis method identifying the hot-spots and cold-spots from geo-spatial data, we highlight and compare the spatial distribution characteristics of the above mentioned metrics. We further compare the local built environment features, such as road density and centrality, building size, count and distance, and proportion of street view elements of greenery, vehicle and human, between the typical hot-spots and cold-spots of above coverage indicators. The aim is to reveal the environmental causes of the potential insufficiency in SVI coverage.

\subsubsection {Exploring the impact of collection interval on SVI coverage}

In the Experiment 3, we aim to investigate the stability of SVI coverage on building facades in terms of different SVI collection intervals. This investigation is based on the hypothesis that smaller collection interval will increase both the completeness and frequency of SVI coverage on the built environment, enhancing the information density, but may not suit best for urban research due to the extra redundancies introduced and the uncertainty in distribution.

This investigation starts with resampling SVI locations at different intervals along the road networks to simulate different SVI collection strategies. For each SVI collection interval and at each H3 level-9 grid, the mean values of \textit{CoC-B} and \textit{FoC-B} are calculated and aggregated, respectively. By observing the variation of the SVI coverage indicators relative to different SVI collection intervals, the study hopes to reveal the potential bias and uncertainty introduced by different SVI collection strategies. 

On this basis, Experiment 3 explores whether there is an optimal SVI collection interval helping improve the reliability of SVI based urban research. Non-linear functions can be fitted to precisely describe the variation of \textit{CoC-B} and \textit{FoC-B} indicators along different intervals, respectively. By analyzing the two fitted functions, especially the speed of indicator increase or decrease relative to the SVI collection interval change, it is hypothesized that we can identify certain intervals which enables sufficient completeness of SVI coverage on built environment information while helping eliminate the unnecessary redundancy and uncertainty.

\subsection {Parameters and settings}
For the standardization of the study and convenience of expression, the Experiment 1 and 2 are carried out based on SVI locations resampled at 50-m interval from the total SVI locations searched. 50-m's searching radius is adopted in the isovist analysis and 50 \% of the building element proportion is applied to decide whether SVI achieve a coverage on building facade on the corresponding directions, thus filtering the isovist analysis results. For the Experiment 3, SVI collection intervals increasing from 10 m to 95 m in 5 m increments are applied to resample the SVI locations. For each interval, the isovist analysis radius is set differently in 30 m, 40 m, and 50 m for comparing the experiment results. The same 50 \% of the building element proportion serves as the filter for isovist analysis results.

\section{Results}

\subsection{Potential bias in the SVI covered building information}

\subsubsection{Coverage completeness distribution by building functions and sizes}
\label{sec:descriptive_summary}

\begin{figure}[htbp]
    \centering
    \begin{subfigure}{1\textwidth}
        \centering
        \includegraphics[scale=0.47]{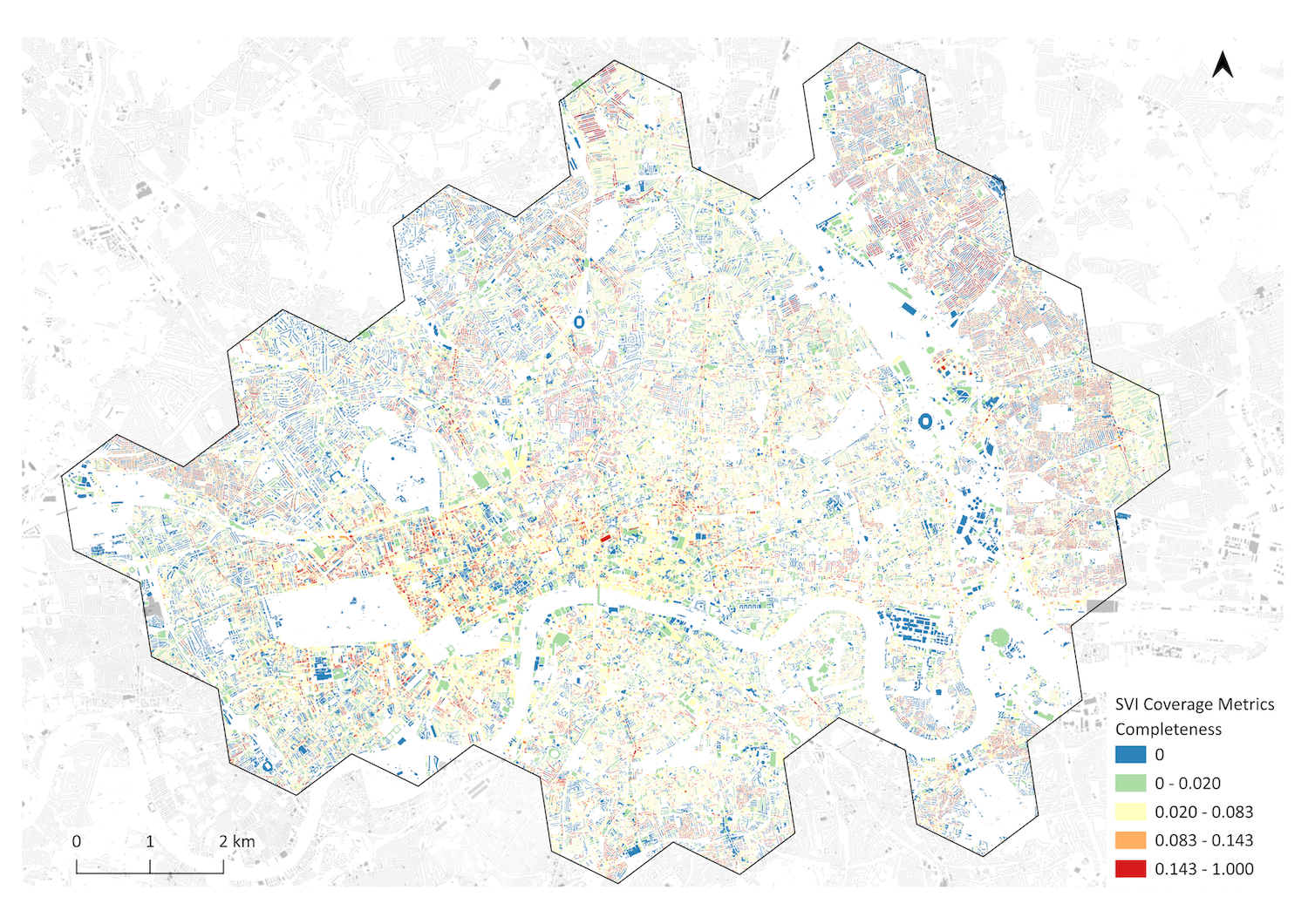}
    \end{subfigure}
    \caption{Map illustrating the distribution of \textit{CoC-B} indicator across the case study area.  }
    \label{fig:Completeness}
\end{figure}

In Experiment 1, the study first applies \textit{CoC-B} indicator to investigate the completeness of SVI coverage on individual buildings.
As a preliminary result, Figure \ref{fig:Completeness} illustrates the spatial distribution of \textit{CoC-B} indicator estimated in the case study area. Notable heterogeneity is observed in this completeness indicator across individual buildings, which varies according to building sizes and locations. Additionally, around 37.6 \% of total buildings are recorded with \textit{CoC-B} equal to 0, indicating that SVI may fail to reach these buildings within the threshold distance, or the buildings are blocked visually by other street elements in SVI. %

\begin{figure}[htbp]
    \centering
    \begin{subfigure}{1\textwidth}
        \centering
        \includegraphics[scale=0.55]{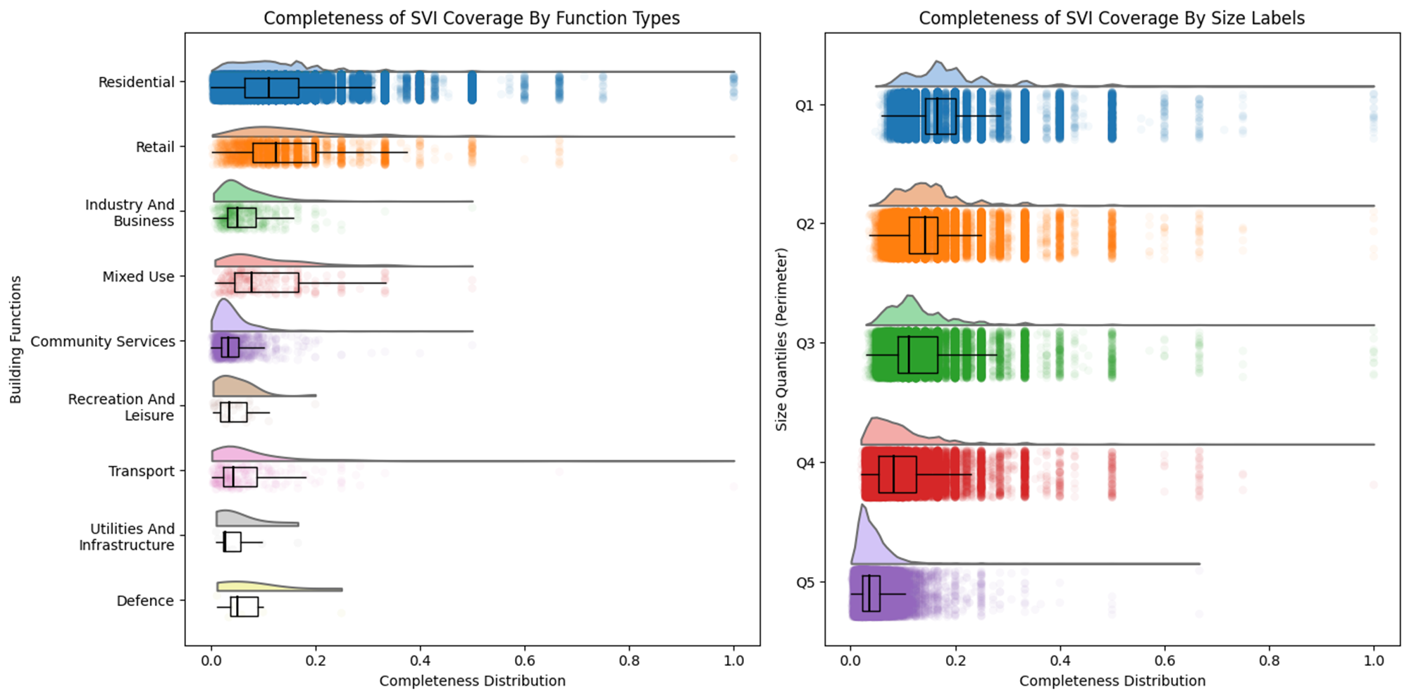}
    \end{subfigure}
    \caption{Boxplots showing the distribution of \textit{CoC-B} for different function types and size quantiles.}
    \label{fig:svi_coverage_box_plots}
\end{figure}

For buildings with \textit{CoC-B} more than 0, Figure \ref{fig:svi_coverage_box_plots} depicts distribution of \textit{CoC-B} values across different building types and sizes. It is found that all building types exhibit 75th percentile values of \textit{CoC-B} indicator below 0.2. This suggests that for the majority of buildings reached by SVI, less than 20 \% of their public facades can be covered. Notably, residential, retail, and mixed-use buildings exhibit significant fluctuations in \textit{CoC-B} distribution, which suggests that the ways SVI achieve coverage on these building types can be more diverse. For residential and retails buildings, they are also featured with higher 25th percentile values of \textit{CoC-B} than other building types, indicating that these building types tend to be covered more completely via SVI. Regarding building size, buildings are re-classified into five groups based on perimeter length, from smallest to largest. It is consistently observed that buildings with longer perimeters exhibit lower \textit{CoC-B} values, suggesting that larger buildings tend to have less complete SVI coverage. Analysis above reveals that the completeness of SVI coverage varies significantly across buildings of different types and sizes, and most of buildings exhibit an insufficient completeness in SVI coverage.

\subsubsection{SVI covered building information and its original distribution}

Beyond completeness, frequency indicators are applied in Experiment 1 to investigate the distribution characteristics of SVI covered building information. The aim is to learn whether SVI tends to under- or over-represent specific building types in covered visual information, relative to their initial proportion in building footprints. 
 
\begin{figure}[htbp]
    \centering
    \begin{subfigure}{1\textwidth}
        \centering
        \includegraphics[scale=0.36]{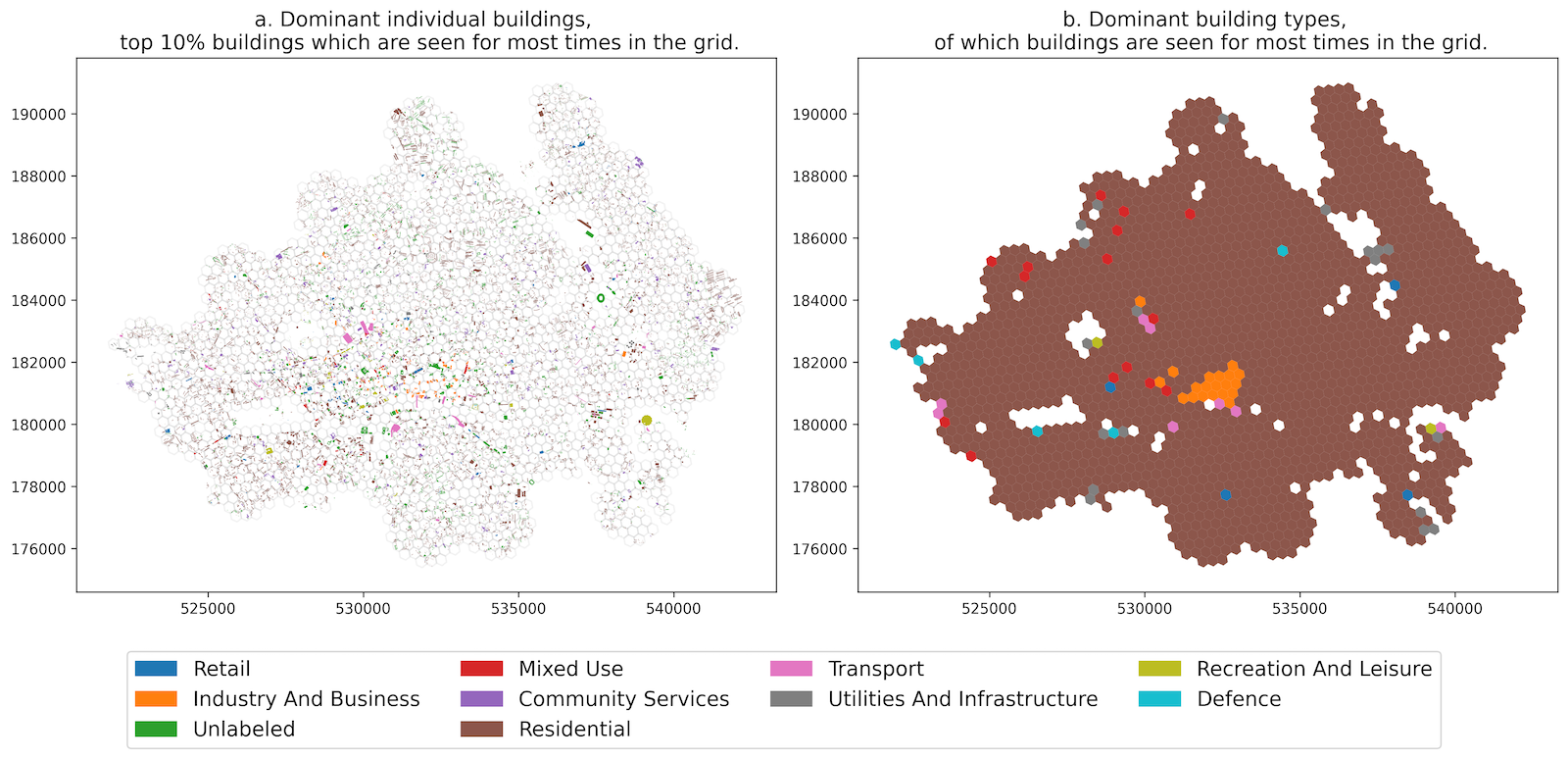}
    \end{subfigure}
    \caption{Mapping the dominant building instance and building types that show highest frequency to view via SVI in each local area.}
    \label{fig:Dominant Building}
\end{figure}

The study first identifies dominant buildings and building types within each H3 level-9 grid, by mapping the top 10 \% of buildings with highest \textit{FoC-B} in the grids, and coloring the grids according to the building types with highest \textit{FoC-A} values, as illustrated in Figure \ref{fig:Dominant Building}. Buildings with traffic, community services, industrial and business functions, along with a multitude of unclassified, large-scale buildings, emerge as the individual buildings more frequently viewed and having a larger visual impact on the local areas. However, when analyzing dominant building types at the grid level, we find that non-residential building types are predominantly visible only within the City of London, the central area of the case study, and in a few isolated grids. Beyond these, residential buildings are the most frequently viewed and serve as general background in the building information covered by most grids in the study area. According to Table \ref{tab:building_function_size} and the method depicted in \ref{sec:sample:appendix_G}, in total there are 62 \% residential buildings covered in SVI, which correspond to 66.2 \% residential population in the study area.

\begin{figure}[htbp]
    \centering
    \begin{subfigure}{1\textwidth}
        \centering
        \includegraphics[scale=0.55]{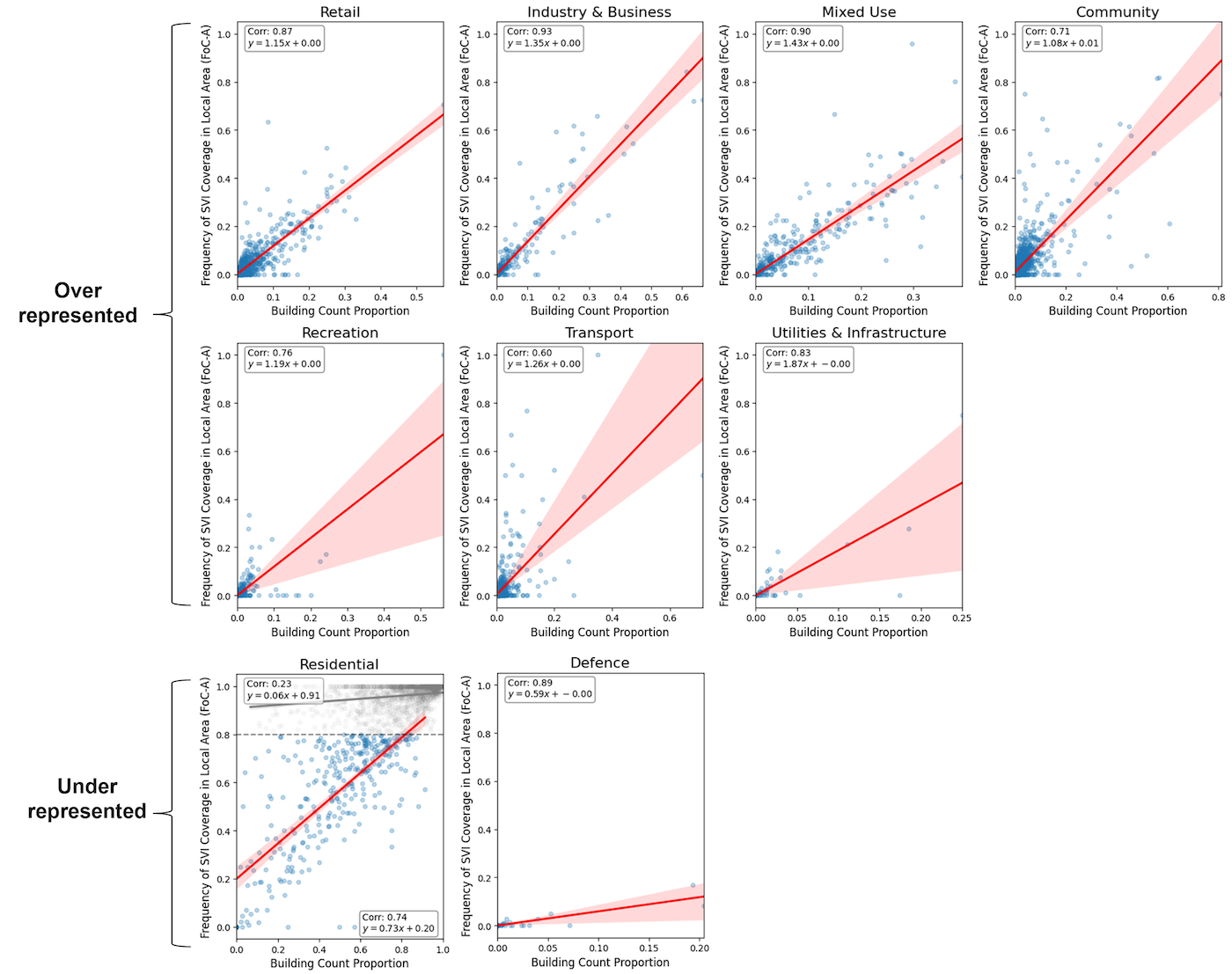}
    \end{subfigure}
    \caption{Scatter plots showing the association between \textit{FoC-A} indicator of a specific building type and count proportion of the building type in H3 level-9 grids. The correlation coefficient and the formula of the fitted trend line are labeled on each subplot.}
    \label{fig:seen_times_proportion_vs_count_proportion}
\end{figure}

The study further explored the linear association between the frequency of building types viewed in SVI (\textit{FoC-A}) and the proportion of building types existing in the grids, as depicted in Figure \ref{fig:seen_times_proportion_vs_count_proportion}. It was found that for building types beyond residential and defence, the regression coefficients of fitted linear equations are all above 1, indicating that SVI tends to over-represent the real presence of these buildings compared to in building footprint data in the study area. 
The pattern is more prominent in industry \& business buildings and mixed use buildings, where the frequency and the proportion variables present highest correlation scores of 0.93 and 0.9, and there are higher coefficients in the linear equations. For residential buildings and buildings with defence usage, SVI tend to under-represent their presence in the built environment. Specifically, for grids with a \textit{FoC-A} of residential buildings less than 0.8, areas with more diverse building function types, an increase of every 1 unit of residential building proportion only explains about a 0.74 unit increase in the frequency of residential buildings viewed. Conversely, for grids with residential buildings dominant in visual information, a limited increase or decrease in residential building proportion has little impact on the frequency with which residential buildings can be viewed via SVI.

The interpretation and analysis on the frequency and completeness indicators above suggests that information of individual buildings are not completely or evenly captured by SVI, and the distribution of SVI covered building information may differ from its distribution in original form and medium. Additionally, the completeness and frequency indicators can serve as effective tools for understanding the potential bias of SVI in representing local environment from a horizontal perspective.

\subsection{Poorly represented neighborhoods in SVI}

\subsubsection{Comparing different SVI coverage measurement methods}

\begin{figure}[htbp]
    \centering
    \begin{subfigure}{1\textwidth}
        \centering
        \includegraphics[scale=0.45]{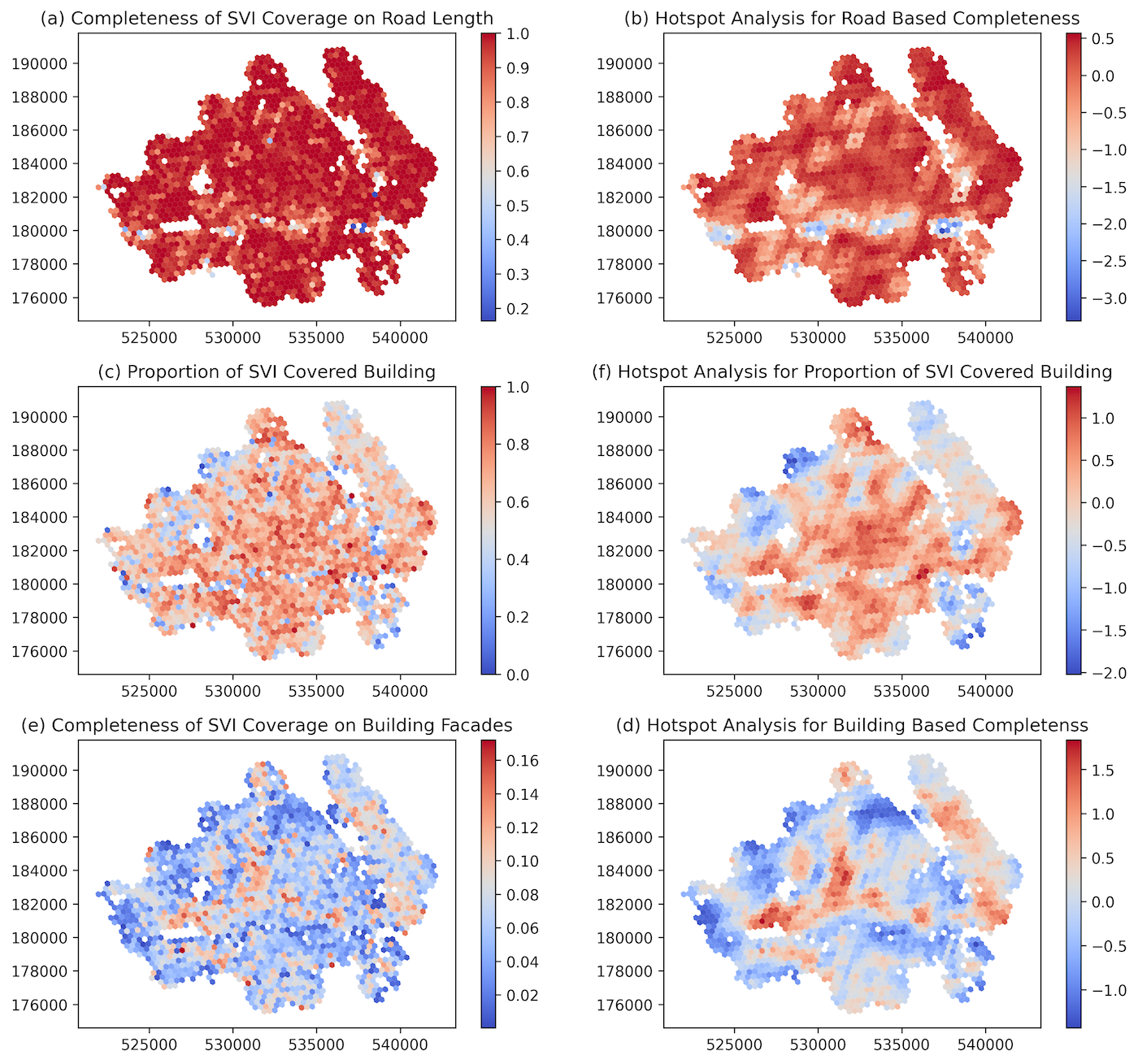}
    \end{subfigure}
    \caption{Distribution of different SVI coverage completeness metrics and the hot-spot analysis results.}
    \label{fig:Different_coverage}
\end{figure}

In Experiment 2 we hope to investigate, whether SVI achieving sufficient coverage in terms of spatial distribution, is equivalent to SVI achieving adequate coverage on building facades. As shown in Figure \ref{fig:Different_coverage}, using the H3 level-9 grid as a basis, we summarized and visualized the completeness of SVI coverage of road lengths within the grid, the proportion of buildings reached by SVI relative to all buildings in the grid (\textit{CoC-A}), and the mean value of coverage completeness for individual building facades within the grid (\textit{CoC-B}). Additionally, we visualized the hot-spot and cold-spot distribution of these indicators using the Getis-Ord Gi* statistics. SVI data was collected from the road networks at equal 50m intervals.

It is found that the three completeness-related indicators show significant differences in their numerical and spatial distribution. The completeness of road length coverage by SVI within each grid generally falls within the high range of 0.8 to 1. Spatially, cold-spots are mainly concentrated along the Thames River. In non-riverbank areas, the degree of SVI coverage for road lengths is relatively uniform. In contrast, the completeness distribution varies significantly when it comes to building facade coverage. Notably, in many grids, only less than 50 \% of buildings can be reached by SVI. Grids with low \textit{CoC-A} values are mostly located in residential-dominated neighborhoods on the periphery of the study area and along the Thames River. Closer to the urban center, the proportion of buildings reached by SVI is relatively higher. 

The completeness of SVI coverage for individual buildings is even lower, with most grids having a mean \textit{CoC-B} indicator value of less than 0.1. Higher values are mainly observed in a continuous band in the central areas of the City, Camden, and Westminster, as well as in eastern residential areas, showing significant local clustering. Low-value grids are primarily found in the western and northern peripheral neighborhoods and along the Thames River.

Based on the above analysis, it can be observed that for most local areas in the case study area, although SVI coverage can achieve sufficient coverage of the road networks, its ability to capture internal block information may be limited, leaving many buildings outside the reach of SVI. Additionally, even if SVI reaches a high proportion of buildings within neighborhoods, the completeness of individual buildings' exposure to SVI may still be significantly lacking. These differences further indicate that SVI collected through equidistant sampling along the road networks may provide misleading information about the built environment. This is particularly important for research evaluating the external environment of buildings at the neighborhood level based on SVI.

\subsubsection{Built environment factors impacting SVI coverage}

\begin{figure}[htbp]
    \centering
    \begin{subfigure}{1\textwidth}
        \centering
        \includegraphics[scale=0.38
]{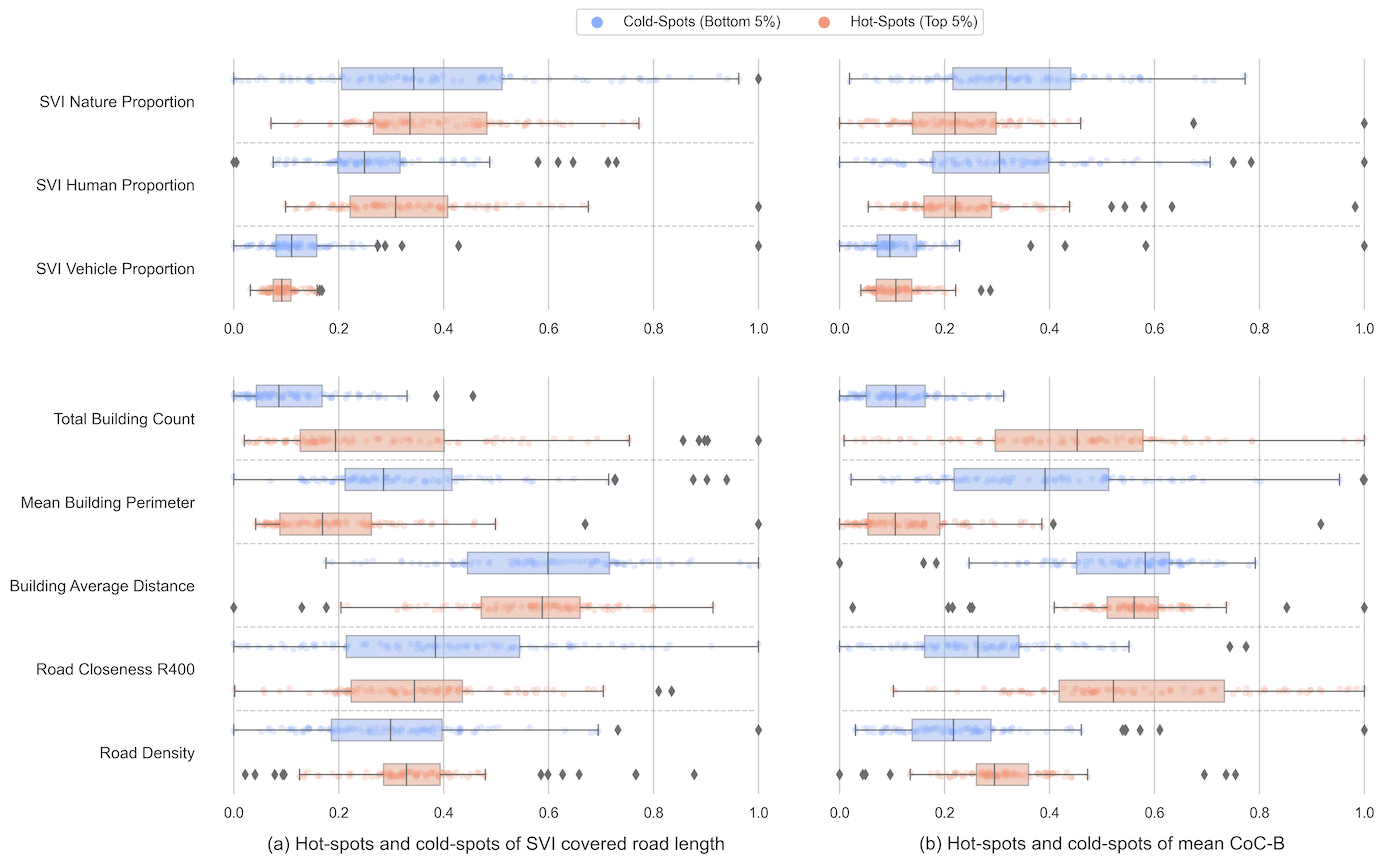}
    \end{subfigure}
    \caption{Distribution of built-environment related indicators for local hot-spots and cold-spots of SVI coverage on road length and building facades.}
    \label{fig:top_bottom_attributes}
\end{figure}

Utilizing Getis-Ord Gi* statistics for completeness indicators on road length and building facades, respectively, we rank the h3 grids and select the top 5 \% and bottom 5 \% as typical hot-spots and cold-spots characterizing the indicators' distribution. Figure \ref{fig:top_bottom_attributes} compares the distribution of
a series of built environment features between the hot-spot and cold-spot grids in each completeness indicator.

For coverage completeness on road length, there is only a limited difference in built environment features between the hot-spot and cold-spot grids. Specifically, the cold-spot grids tend to have relatively lower building counts, lower road density, and longer building perimeters compared to the hot spots. These characteristics align with their consistent distribution along the riverbank, where large buildings cluster and road access is limited. In contrast, significant differences can be observed between hot-spots and cold-spots for the completeness indicator on building facades. Cold-spots show lower road density, lower building counts, and longer building perimeters. Additionally, these cold-spots feature higher proportions of natural and human elements in SVI, and significantly lower closeness values. The former indicates that greens and humans rather than vehicles may play more positive roles in blocking building elements within image space of SVI. While the latter suggests that grids far from local urban centers or high streets, and deeper within neighborhoods are prone to having poorer coverage of building facades.

\subsection{Impact of collection interval on SVI coverage}
\subsubsection{Robustness of SVI coverage across different SVI collection intervals}

\begin{figure}[htbp]
    \centering
    \begin{subfigure}{1\textwidth}
        \centering
        \includegraphics[scale=0.43]{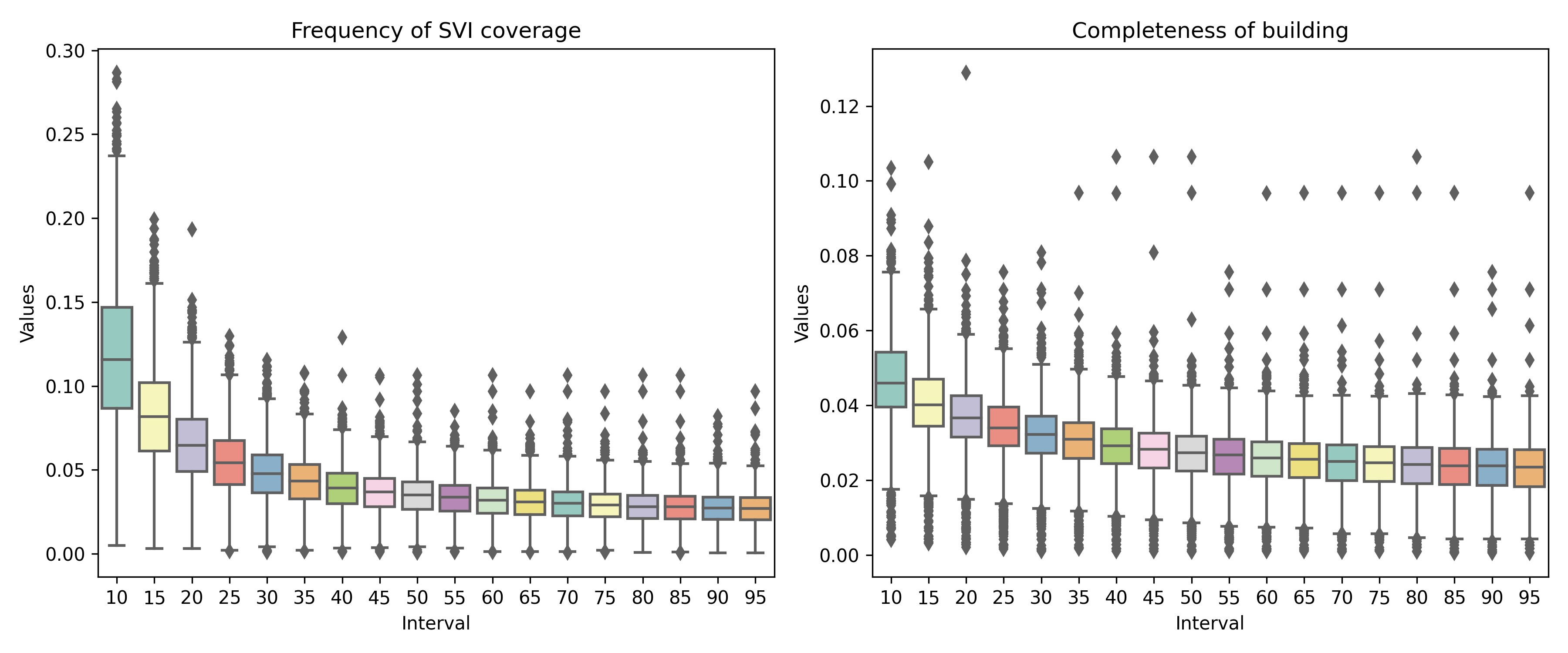}
    \end{subfigure}
    \caption{Box plots drawn based on  SVI coverage indicators of different local grids and at different SVI collection intervals. }
    \label{fig:boxplots_of_SVI_Coverage}
\end{figure}

In Experiment 3, we explore how different SVI collection intervals impact the values and distributions of SVI coverage indicators. Figure \ref{fig:boxplots_of_SVI_Coverage} presents the distribution of \textit{CoC-B} and \textit{FoC-B}, the building-level completeness and frequency SVI coverage indicators, with SVI collection intervals increasing from 10 m to 95 m in 5 m increments. The observations are mean values of indicators aggregated at h3 level-9 grids across the case study area. It can be identified that when applying smaller SVI collection intervals, the \textit{FoC-B} indicators of SVI coverage show larger differences between different grids, while the differences decrease significantly when larger collection intervals are employed. In comparison, the distribution differences of the \textit{CoC-B} indicators across grids are less affected by the SVI collection interval. In addition, the decreasing of mean frequency across grids along with the increasing of SVI collection interval tend to be faster than the decreasing of mean completeness. The results indicate that the completeness indicators show stronger robustness to changes in SVI collection interval, in terms of spatial distribution and numerical values. In contrast, the frequency indicators can be sensitive to lower SVI collection intervals and higher SVI density.

\subsubsection{Optimal SVI collection intervals for frequency and completeness}
Beyond the differences in robustness, 
the study attempts to explore whether there exists an optimal SVI collection interval that achieves higher SVI coverage completeness while avoiding unnecessary high SVI coverage frequency, taking the speed difference in the decline of frequency and completeness indicators alongside rising SVI collection interval as an entry point.

\begin{figure}[htbp]
    \centering
    \begin{subfigure}{1\textwidth}
        \centering
        \includegraphics[scale=0.43]{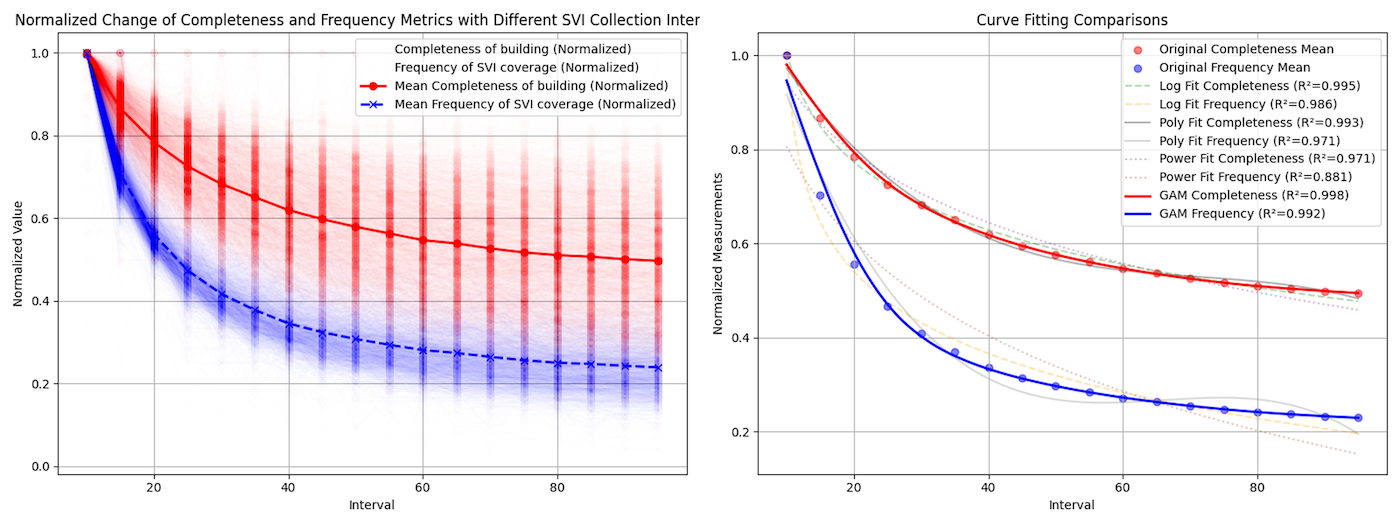}
    \end{subfigure}
    \caption{Left: Normalized values of \textit{CoC-B} and \textit{FoC-B} of different local grids and at different SVI collection intervals, with the mean values highlighted.
    Right: A comparison of curve functions fitted with generalized additive model and other modeling methods, based on the mean values. }
    \label{fig:SVI_Coverage_curves}
\end{figure}

The study first normalizes the values of completeness and frequency in each H3 level-9 grid for each SVI collection interval based on the corresponding values at the minimum SVI collection interval within the grid. As shown in Figure \ref{fig:SVI_Coverage_curves}, plotting the normalized values into the same quadrant reveals the trend of both indicators shrinking relative to their maximum values as the SVI collection interval increases. It is observed that the shrinking speed of frequency is significantly higher than that of completeness at lower stages of the SVI collection interval, whether for a single grid or on an average level of grids. With the increase in the SVI collection interval, the rate of shrinkage for both indicators slows down and converges.

Based on this observation, it can be hypothesized that there exists a specific interval threshold where the decrease speeds of completeness and frequency are equal. Below this interval, the decrease speed of frequency exceeds that of completeness, indicating that the reduction in redundancy in the environment information captured by SVI is faster than the decrease in completeness, and it is economical to continue reducing the SVI collection interval. Above this interval, the decrease speed of frequency is lower than that of completeness, indicating that further reducing the SVI collection interval would relatively more severely affect the completeness of the environmental information captured by SVI. Collecting SVI at this interval threshold can be seen as an optimal strategy to balance the completeness and redundancy of the information captured by SVI.

To precisely identify the optimal interval, the study attempts to fit functions to the normalized values of completeness and frequency at different SVI collection intervals in each grid and identify derivative curves based on them. The interval corresponding to the intersection point of the two derivative curves is considered the optimal interval. Polynomial function, power function, logarithm function, and generalized additive model (GAM) are applied to fit curves with the observations, respectively. Among them, the GAM model is identified as the best model to capture both the global decreasing trend of completeness and frequency indicators (the highest R2) and their subtle changes in local regions. Figure \ref{fig:SVI_Coverage_curves} shows that the match between original observations and the GAM-fitted curves is significantly better than that for other curves. 

\begin{figure}[htbp]
    \centering
    \begin{subfigure}{1\textwidth}
        \centering
        \includegraphics[scale=0.37]{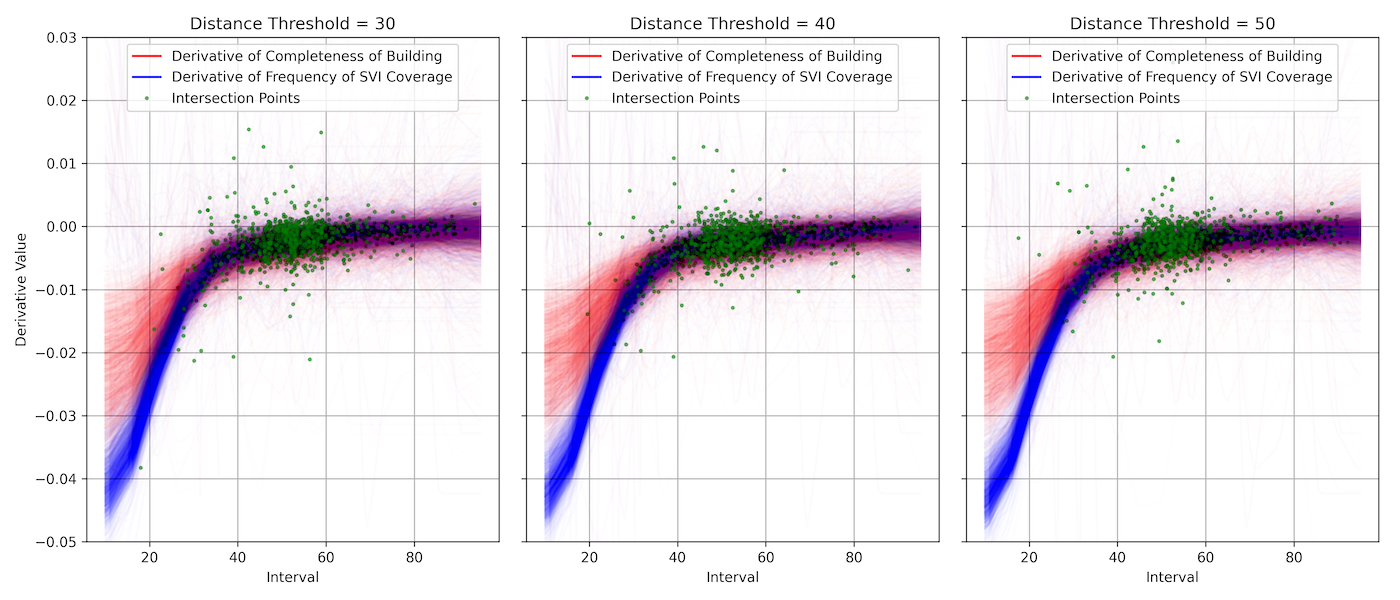}
    \end{subfigure}
    \caption{Derivative curves of completeness and frequency across different local grids and under different isovist analysis radii, and the intersection points between the paired curves.}
    \label{fig:derivative_curves}
\end{figure}

\begin{figure}[htbp]
    \centering
    \begin{subfigure}{1\textwidth}
        \centering
        \includegraphics[scale=0.33]{
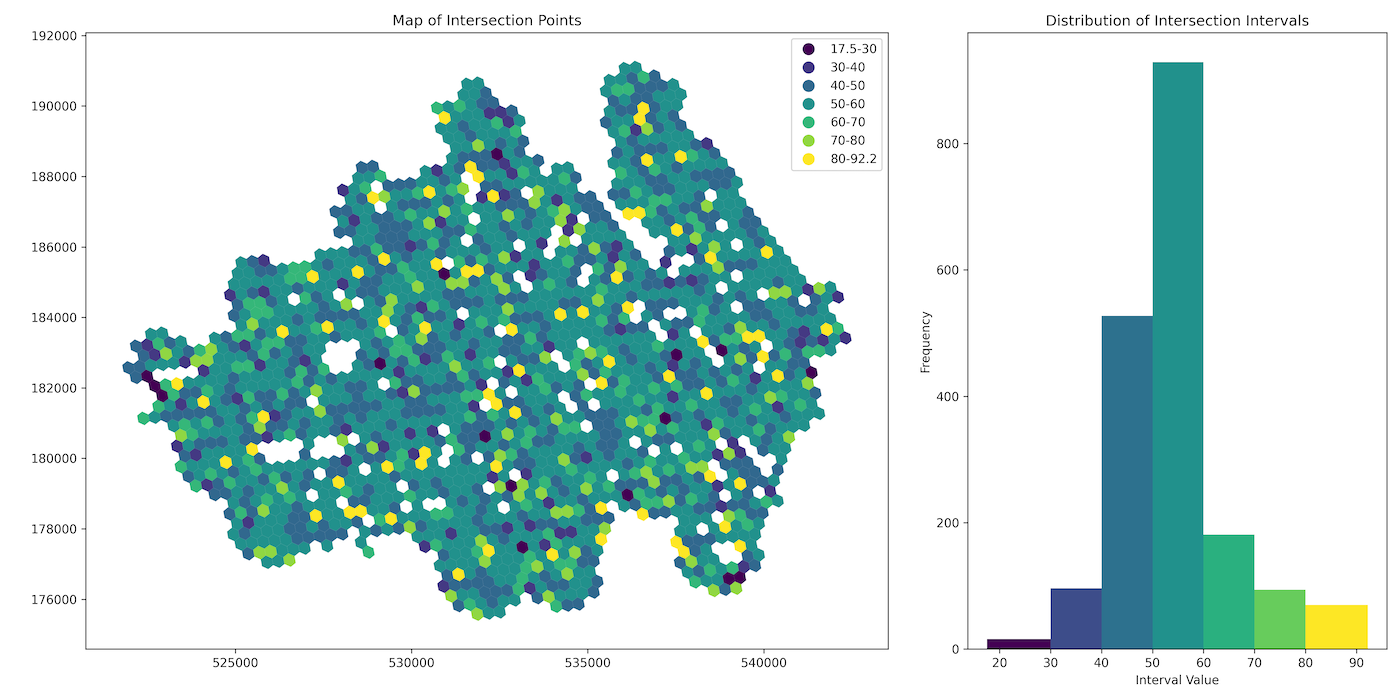}
    \end{subfigure}
    \caption{Spatial and data distribution of optimal SVI collection intervals detected from each H3 level-9 grid in the case study area.}
    \label{fig:interval_distribution}
\end{figure}

Figure \ref{fig:derivative_curves} presents the distribution of individual derivative curves fitted based on each grid's completeness and frequency observations, and the intersection points from the paired curves. Considering that the distance thresholds applied in isovist analysis may also have an impact on the distribution of SVI coverage indicators, the figure further compares the distribution of curves and intersection points across isovist analysis distance thresholds of 30 m, 40 m and 50 m.  In general, the derivative curves fitted present significant trends for intersecting after SVI collection interval of around 30 m. More intersection points are densely distributed between 50 m - 60 m interval. The increasing of distance threshold of isovist analysis shows little impact on the location distribution of intersection points.

Taking analysis results under the isovist analysis thresholds of 50 m as an example, Figure \ref{fig:interval_distribution} further plots the spatial and data distribution of detected optimal SVI collection interval for each H3 level-9 grid. It is found that grids with optimal SVI collection intervals around 50 m - 60 m present a relatively even distribution across grids in the study area, without showing significant spatial clustering.  The analysis results above demonstrate that, it's possible to employ a consistent and stable optimal interval at a large spatial scale for SVI data collection.

\section{Discussion}

\subsection{Uneven SVI coverage at element level}

SVI is typically collected at equal intervals along roads to achieve even mapping and reliable representation of the urban environment.  However, Experiment 1 reveals that SVI collected in this manner does not translate into uniform coverage of buildings of different types and sizes, particularly in terms of facade completeness. Residential and retail buildings, and buildings with smaller sizes tend to have more complete facade coverage in SVI. Moreover, the distribution of SVI-covered buildings does not align with the actual distribution of buildings in footprint. Specifically, SVI tends to significantly over-represent building types such as mixed use, industry \& business, and transport. Community and retail buildings are slightly over-represented, while residential buildings, especially in suburban areas, are generally under-represented.

The practical impact of these disparities depends on the specific research purpose. If SVI is used for urban perception studies focused only on visible elements or experiences at SVI locations, uneven coverage may not affect reliability. However, using limited SVI data to characterize an entire region may over-represent frequently covered elements and under-represent others. This implicit prioritization of certain environmental elements in SVI can be considered a form of spatial weighting, which has been largely overlooked in previous urban studies and adds uncertainty to interpretations of SVI-based urban perception studies. Nevertheless, the element-level coverage estimation method proposed in this study helps address this gap by providing a clearer understanding of how different elements and instances are represented in SVI data.

\subsection{Element-level coverage as a new Dimension for SVI data quality}
In Experiment 2, the study compares the proposed element-level SVI coverage estimation method, which considers both SVI locations and environmental obstructions, with traditional methods that only consider SVI locations. It is found that, even in regions where traditional methods aim to achieve complete SVI coverage, \textit{CoC-A} and \textit{CoC-B} indicators are often low, reflecting a lack of building instances reached by SVI and incomplete facade coverage. The spatial distribution of SVI building facade coverage shows strong spatial auto-correlation, with hot-spots and cold-spots linked to factors like building size, road network centrality, density, and obstacles like greenery and human. The results further explain the implicit prioritization of certain environmental elements in SVI. It is not determined by the subjective intentions of researchers, but rather by the complex spatial configuration and obstruction relations shaped collectively by the roads, buildings and other environmental elements.

Based on this comparison, we propose introducing the extent of SVI coverage on urban environmental elements as a novel dimension in SVI data quality assessments. This dimension addresses the impact of environmental obstructions on SVI usage. Moreover, it highlights SVI's capacity to reveal urban information in the horizontal dimension, distinguishing SVI from other data forms such as building footprints or satellite imagery, which conventionally provide vertical perspectives. In the data quality evaluation framework by \citet{hou_comprehensive_2022}, environmental obstruction is treated as part of image quality issues. However, we argue that obstruction-related problems should be considered as an external dimension. Obstruction, such as buildings blocked by vegetation or other buildings, may not occur randomly but instead follow patterns related to spatial and socioeconomic factors, such as residential density.

The method proposed in this study was tested using Google Street View and commercial SVI in general, but it can also be applied to volunteered street view imagery (VSVI) sources, only if key SVI metadata is available. For panoramic images, this includes image heading and coordinates, while for perspective images, additional information such as field of view (FOV) and rotation is required. If SVI data quality is low or reference data, such as building footprints, is incomplete, the proposed method can be supplemented with additional approaches. In such cases, the relationship between SVI coverage extent and built environment features can be modeled in areas with sufficient data, and this model can then be used to infer the bias risks of SVI coverage in regions lacking adequate data, guiding the data collection and utilization strategies. The global OSM building completeness dataset created by \cite{herfort_spatio-temporal_2023} serves as a relevant example of this approach. Overall, our element-level SVI coverage estimation framework demonstrates significant potential to enhance the reliability of SVI in urban analysis and perception studies, with broad applications across related fields.

\subsection{Does an optimal SVI collection interval really exist?}

In this study, we also test the impact of different SVI collection intervals on the element-level SVI coverage. 
The building-level completeness and frequency indicators, \textit{CoC-B} and \textit{FoC-B}, act as a pair of complementary indicators to determine whether the SVI coverage of building information is sufficient and whether there is potential redundancy. By calculating these indicators across various intervals, we found that a 10 m sampling interval significantly increases the average completeness and frequency of SVI coverage for element instances compared to a 50 m interval, which can notably influence subsequent analysis based on SVI. This result aligns with the findings of \citet{kim_decoding_2021}.

Our study goes further by revealing that \textit{CoC-B} and \textit{FoC-B} decrease non-linearly as the SVI sampling interval increases, but at different rates. We identified a critical interval: below this threshold, the frequency indicator (\textit{FoC-B}) decreases faster than the completeness indicator (\textit{CoC-B}); above it, the reverse occurs. This suggests that SVI collected near this critical interval threshold balances higher building coverage completeness with lower redundancy, maximizing cost-effectiveness. Further analysis confirmed this critical interval, typically ranging between 50 m and 60 m, is consistently distributed across most local grids, supporting current common practices and providing valuable guidance for future SVI-based urban research.

Nevertheless, due to the diverse use cases of SVI, a universal SVI sampling interval may not exist for all applications. The intervals identified in this study are mainly suited for SVI as a comprehensive visual representation of the environment, particularly for applications such as spatial perception and experience studies. However, for mapping specific environmental elements, such as buildings or trees, smaller intervals and denser SVI sampling—while introducing more redundancy—can effectively enhance coverage completeness and reduce data gaps.

\section{Conclusion}
This paper introduces a comprehensive workflow to estimate the element-level coverage of SVI, taking urban building facades as an example, and has further explored the potential of coverage extent as novel indicators in validating the usage of SVI in urban and spatial analytics. Our study shows that, despite dense availability on urban road networks, SVI only reaches 62.4 \% of buildings in the case study area. The completeness of SVI coverage on building facades remains low, averaging 12.4 \%, with large differences depending on building types and sizes. Further, besides revealing data gaps and inconsistent coverage, our results indicate questionable representativeness of SVI --- there are biases in the information collected from SVI, with some instances being over- or under-represented. These biases can impact the integrity of urban studies relying on SVI. For example, if one is using SVI to infer the share of certain building types in streets, with some instances being omitted more often than others in the images collected from cars on roads, the result would not be entirely accurate.

A potential range of optimal SVI sampling intervals, 50--60 m, is identified to help achieve a better application of SVI data. SVI data has been used widely across multiple disciplines, but data quality and integrity have not been given adequate attention. Regarding both the infrastructural and human aspect, for the first time, we reveal at a very high resolution and large-scale, the reach and usability of SVI for urban sensing and mapping. Our study argues that the element-level coverage of SVI, with respect to building, greenery, and other useful street view elements and visual information, should be included as a new dimension for SVI data quality assessment.

Nonetheless, the study has some limitations that offer opportunities for future work. We believe that the findings and results will depend on the particular context --- our study focuses on a particular use case (mapping buildings) in a particular location (London), so further investigations are necessary. Next, to balance detail and scalability, the study does not fully account for variations in building heights within the SVI coverage estimation workflow. However, it is possible that even if one building is blocked by another building according to the 2D isovist analysis based on building footprints, part of the blocked building may still be visible from SVI due to differences in building heights. Incorporating 3D urban data, such as Digital Surface Models (DSM) and LiDAR point clouds, along with 3D isovist analysis, presents opportunities to enhance analytic precision in future developments. Another concern is the limitation of SVI in capturing temporal variation, especially in measuring dynamic street objects, such as pedestrians and vehicles, and in reflecting the seasonal change of vegetation \citep{liu_clarity_2024,yan_evaluating_2023}. 
The limitation introduces disturbance when incorporating environmental obstructions in the element-level SVI coverage estimation.

\section*{Acknowledgments}
We thank our colleagues at the NUS Urban Analytics Lab for the discussions. 
The first author is supported by the National University of Singapore under the President’s Graduate Fellowship.
This research is part of the project Large-scale 3D Geospatial Data for Urban Analytics, which is supported by the National University of Singapore under the Start Up Grant R-295-000-171-133.

\section*{Author contributions}

\textbf{Zicheng Fan:}
Conceptualization;
Methodology; 
Software;
Validation;
Formal analysis;
Investigation;
Data Curation;
Writing - Original Draft;
Visualization.

\textbf{Chen-Chieh Feng:}
Conceptualization;
Writing - Review \& Editing;
Supervision;

\textbf{Filip Biljecki:}
Conceptualization;
Investigation;
Methodology;
Writing - Review \& Editing;
Visualization;
Supervision;
Project administration;
Funding acquisition.

\section*{Declaration of generative AI and AI-assisted technologies in the writing process}

During the preparation of this work the authors used ChatGPT in order to proofread the text. After using this tool, the authors reviewed and edited the content as needed and take full responsibility for the content of the publication.

\newpage
\appendix
\section{An Overview of Studies on Mapping and Sensing Urban Environmental Elements Using SVI Data}
\label{sec:sample:appendix_A}

\begin{table}[h!]
\scriptsize
\caption{Overview of latest studies on mapping and sensing urban environmental elements using SVI data, with a focus on individual elements such as roads, buildings, greenery, and the use of combined elements.}
\centering
\begin{tabular}{p{1.7cm} p{4.5cm} p{5.3cm}}
\toprule
\textbf{Elements} & \textbf{Use Cases} & \textbf{References} \\ 
\midrule

\multirow{4}{*}{\textbf{Road}} 
& Damage detection & \citet{ren_yolov5s-m_2023} \\ \cline{2-3}
& Sidewalk mapping & \makecell[l]{\citet{hamim_mapping_2024}, \\ \citet{de_mesquita_street_2024}, \\ \citet{ning_sidewalk_2022}} \\ \cline{2-3}  
& Wheelchair usage & \citet{ning_converting_2022} \\ 
\midrule

\multirow{6}{*}{\textbf{Building}} 
& Perception on exteriors & \citet{liang_evaluating_2024} \\ \cline{2-3}
& Building typology & \citet{gonzalez_automatic_2020} \\ \cline{2-3}
& Building height & \citet{yan_estimation_2022} \\ \cline{2-3}
& Age and style & \citet{sun_understanding_2022} \\ \cline{2-3}
& Building material & \citet{raghu_towards_2023} \\ \cline{2-3}
& Building color & \citet{zhou_evaluating_2023} \\ \cline{2-3}
& Building usage & \citet{ramalingam_automatizing_2023,Ramalingam2025} \\ \cline{2-3}
& Energy efficiency & \citet{mayer_estimating_2023} \\ \cline{2-3}
& Seismic vulnerability & \makecell[l]{\citet{ruggieri_machine-learning_2021}, \\ \citet{aravena_pelizari_automated_2021}} \\ \cline{2-3}
& Flood risk & \citet{xing_flood_2023} \\  \cline{2-3}
& Abandoned houses & \citet{zou_mapping_2022} \\ 
\midrule

\multirow{3}{*}{\textbf{Greenery}}
& Green View Index distribution & \citet{zhang_spatial_2024} \\ \cline{2-3}
& Greenery visibility & \citet{sanchez_accessing_2024} \\ \cline{2-3}
& Trees mapping & \makecell[l]{\citet{liu_establishing_2023}, \\ \citet{lumnitz_mapping_2021}} \\ \cline{2-3}
& Tree species detection & \citet{choi_automatic_2022} \\ \cline{2-3}
& Crop types detection & \citet{yan_exploring_2021} \\ \cline{2-3}
& Street forest & \citet{liang_assessment_2023} \\ 
\midrule

& Impression and perceptions & \makecell[l]{\citet{ogawa_evaluating_2024}, \citet{dong_assessing_2023}, \\ \citet{inoue_landscape_2022}} \\ \cline{2-3} 
& Road safety and accidents & \makecell[l]{\citet{ye_unpacking_2024}, \citet{yu_can_2024}} \\ \cline{2-3}
& Gentrification & \citet{thackway_implementing_2023} \\ \cline{2-3}
& Potential of urban renewal & \citet{he_extracting_2023} \\ \cline{2-3}
\textbf{Combined} & Poverty & \citet{yuan_using_2023} \\ \cline{2-3}
\textbf{Elements} & Spatial quality & \citet{rui_quantifying_2023} \\ \cline{2-3}
& Physical disorder & \citet{chen_measuring_2023} \\ \cline{2-3}
& Temporal evolution & \citet{liang_revealing_2023} \\ \cline{2-3}
& Violent crime, travel and & \multirow{2}{*}{\citet{fan_urban_2023}} \\
& health behavior & \\ 
\bottomrule

\end{tabular}
\label{table:urban_features}
\end{table}

\newpage
\section{Building Threshold Calculation}
\label{sec:sample:appendix_B}
With Formula \ref{formula:proportion}, we calculate the proportion of general building elements (buildings, walls, fences) with respect to all non-void, non-flat, and non-sky elements. The proportion serves as threshold to examine whether in the line of sight direction, building facades are visible effectively from SVI location and to quantify the potential impact of environmental obstructions such as vehicles, trees, and pedestrians on SVI coverage. Label id and colormap from the Cityscapes segmentation benchmark is applied \citep{cordts_cityscapes_2016}. The proportion calculated serves as a threshold to help filter the lines of sight that reach building facades for each SVI.

\begin{equation}
    P_{\text{building}} = \frac{\sum_{i \in \mathcal{B}} A_i}{\sum_{j \in (\mathcal{E} \setminus \{\text{void}, \text{flat}, \text{sky}\})} A_j}
    \label{formula:proportion}
\end{equation}

\noindent where:
\begin{itemize}
    \item $\mathcal{B}$: The set of selected building elements, which includes: Building ($id = 11$), Wall ($id = 12$), Fence ($id = 13$).
    \item $A_i$: The area of element $i$.
    \item $\mathcal{E} \setminus \{\text{void}, \text{flat}, \text{sky}\}$: The set of all elements excluding the following categories:
    \begin{itemize}
        \item \textbf{Void}: Unlabeled ($id = 0$), ego vehicle ($id = 1$), rectification border ($id = 2$), out of ROI ($id = 3$), static ($id = 4$), dynamic ($id = 5$), ground ($id = 6$).
        \item \textbf{Flat}: Road ($id = 7$), sidewalk ($id = 8$), parking ($id = 9$), rail track ($id = 10$).
        \item \textbf{Sky}: Sky ($id = 23$).
    \end{itemize}
\end{itemize}

\newpage
\section{Summary of the Completeness and Frequency Indicators}
\label{sec:sample:appendix_C}
The study proposes an indicator system to describe the extent of SVI coverage on building element. Below is a table summarizing the SVI coverage indicators proposed and applied in the study.

\begin{table}[ht]
    \centering
    \small
    \caption{Dimensions, Metrics, and Description of SVI Coverage}
    \begin{tabular}{ l p{4cm} p{6cm} }
        \toprule
        \textbf{Dimensions} & \textbf{Metrics} & \textbf{Description} \\ \midrule
        \multirow{2}{*}{Building Level} & Completeness of SVI Coverage for Individual Building (CoC-B) & Proportion of SVI-covered sampling points relative to the total available sampling points for a single building. \\ \cline{2-3} 
        & Frequency of SVI Coverage for Individual Building (FoC-B) & Number of occurrences SVI achieves coverage around a single building, adjusted for building perimeter. \\ \midrule
        \multirow{2}{*}{Area Level} & Completeness of SVI Coverage on Buildings in Local Area (CoC-A) & Proportion of SVI-covered buildings relative to the total number of buildings in a local area. \\ \cline{2-3} 
        & Frequency of SVI Coverage on Buildings in Local Area (FoC-A) & Sum of SVI coverage frequency for specific building types (not adjusted), relative to the total SVI coverage frequency for all buildings in the local area. \\ \toprule
    \end{tabular}
    
    \label{tab:svi_coverage}
\end{table}

\newpage
\section{A Re-classification of OSM Building Types}
\label{sec:sample:appendix_D}

The study adopts building-level land use data from the Colouring Cities Research Programme (CCRP), an open building data project managed by The Alan Turing Institute, as the basis for categorizing building footprints into different types. Building type information from OSM is also utilized as a complement for building footprints whose land use information is missing.

\begin{table}[htbp]
    \centering
    \scriptsize
    \caption{Corresponding between CCRP land use types and OSM building types. The study gives priority to the CCRP land use data corresponding to the building footprint. In cases where land use data is missing, if the OSM building type information is not empty, then convert the OSM classification into the corresponding land use categories. }
    \begin{tabular}{lp{9cm}}
    \toprule
     \textbf{CCRP Land Use Types}& \textbf{OSM Building Type Labels}\\
    \midrule
     Residential & `apartments', `flats', `house', `terrace', `detached', `semidetached\_house', `dormitory', `hall\_of\_residence', `cottage', `bungalow', `terrace\_house', `council\_flats', `farm\_auxiliary', `farm', `houseboat', `stable', `cabin', `terraced\_house', `Nursery,\_School', `yes;dormitory' \\
    \midrule
     Mixed Use & `yes, office, shop, r', `apartments;residenti', `apartments;yes', `commercial;detached', `retail;yes' \\
    \midrule
     Industry\ and Business & `office', `data\_center', `commercial', `warehouse', `industrial', `light\_industrial', `factory', `manufacture', `office;yes', `telecommunication', `business', `artists\_studio' \\
    \midrule
     Community\ Services & `church', `university', `school', `government', `public', `hospital', `college', `Community\_Building', `kindergarten', `memorial',  `student\_residence', `gatehouse', `cafe', `greenhouse', `monument', `pavilion', `palace', `mosque', `synagogue',  `police\_station', `religious', `clock\_tower',  `village\_hall', `conservatory', `chapel' \\
    \midrule
     Retail & `retail', `pub', `kiosk', `stall',  `bar', `shop' \\
    \midrule
     Transport & `train\_station', `transportation', `ship', `boat', `bridge', `railway\_arch', `railway', `bus', `viaduct', `tunnel\_mouth', `tunnel\_entrance', `bus\_garage' \\
    \midrule
     Recreation\ and Leisure & `civic', `hall', `ruins', `stadium', `recreational', `gallery', `theatre', `cinema', `museum', `sports\_centre', `sports\_hall', `swimming\_pool', `parking', `yes;public;sports\_ce' \\
    \midrule
     Utilities\ and Infrastructure & `service', `construction', `roof', `vent\_shaft', `air\_shaft', `ventilation\_shaft', `electricity', `substation', `gasometer', `air\_vent', `tunnel\_shaft', `water' \\
    \midrule
     Vacant\ and Derelict & `vacant', `disused\_station', `abandoned', `ruins' \\
    \midrule
     Defence & `guardhouse', `bunker', `barracks' \\
    \midrule
     Unclassified & `None', `yes', `no', `multiple', `part' \\
    \bottomrule
    \end{tabular}
    \label{tab:building_classification}
\end{table}

\newpage
\section{Completeness of SVI Coverage on Road Networks}
\label{sec:sample:appendix_E}

The completeness of SVI coverage along the road networks is calculated according to the Formula \ref{formula:SVI_road_completeness}. Buffering analysis with a radius of 50 m is carried out based on each SVI location. Then we calculate the proportion of buffered road length with respect to the total road length within each local area.

\begin{equation}
    SVI Coverage_{\text{road}} = \frac{\sum_{r \in \mathcal{R}} L_r}{\sum_{t \in \mathcal{T}} L_t}
    \label{formula:SVI_road_completeness}
\end{equation}

\noindent where:
\begin{itemize}
    \item $\mathcal{R}$: The set of road segments covered within a 50 m buffer around SVI locations.
    \item $L_r$: The length of road segment $r$ within the buffered area.
    \item $\mathcal{T}$: The total set of road segments in the local area being analyzed.
    \item $L_t$: The total length of road segment $t$ in the local area.
\end{itemize}

\newpage
\section{Completeness Distribution By Building Functions and Sizes}
\label{sec:sample:appendix_F}

For buildings with different type and size labels, Table \ref{tab:building_function_size} compares how the proportions of SVI covered buildings vary. Approximately 62.4 \% of the total buildings are visible from at least one SVI location within the case study area.  

\begin{table}
    \scriptsize
    \centering
    \caption{The table illustrates the proportion of buildings in different types and sizes that have at least one sampling point visible from SVI locations and the related \textit{CoC-B} values.
    The results are based on SVI locations resampled with a 10 m interval and the isovist analysis carried out based on a radius of 50m. The visible building sampling points are filtered with a threshold that the building element proportion visible along the observing direction should be over 50 \% }
    \begin{tabular}{l r r r }
        \toprule
        \makecell[l]{\textbf{Building} \\ \textbf{Type}} & \makecell[l]{\textbf{Proportion}\\ \textbf{of SVI Covered} \\ \textbf{Building}} & \makecell[l]{\textbf{Mean} \\ \textbf{Completeness} \\ \textbf{(All)}} & \makecell[l]{\textbf{Mean}  \\ \textbf{Completeness} \\ \textbf{(SVI Covered)}} \\ 
        \midrule
        \makecell[l]{\textbf{Building Function}} \\
        \midrule
         Residential & 0.619621 & 0.077682 & 0.125371 \\ 
         Retail & 0.698083 & 0.105884 & 0.151678 \\ 
         \makecell[l]{Industry And Business} & 0.796421 & 0.052092 & 0.065407 \\ 
         \makecell[l]{Mixed Use} & 0.829787 & 0.089946 & 0.108397 \\ 
         \makecell[l]{Community Services} & 0.642818 & 0.027731 & 0.043140 \\ 
         \makecell[l]{Recreation And Leisure} & 0.507937 & 0.023605 & 0.046472 \\
         Transport & 0.675287 & 0.049409 & 0.073167 \\ 
         \makecell[l]{Utilities  And \\ Infrastructure} & 0.421053 & 0.021906 & 0.050206 \\ 
         Defence & 0.300000 & 0.029499 & 0.083163 \\ 
         \makecell[l]{Vacant And Derelict} & 0.500000 & 0.016393 & 0.032787 \\ 
         \makecell[l]{Unclassified, \\ presumed non-residential} & 0.820789 & 0.041508 & 0.050571 \\ 
         Unlabeled & 0.595276 & 0.072217 & 0.121317 \\ 
        \midrule
        \makecell[l]{\textbf{Building Size} \textbf{(Perimeter Ranking)}} \\
        \midrule
         \makecell[l]{Q1 (0-20 \%)} & 0.532637 & 0.104767 & 0.196695 \\ 
         \makecell[l]{Q2 (20 \%-40 \%)} & 0.594441 & 0.095654 & 0.160915 \\ 
         \makecell[l]{Q3 (40 \%-60 \%)} & 0.615115 & 0.084116 & 0.136748 \\ 
         \makecell[l]{Q4 (60 \%-80 \%)} & 0.626316 & 0.067932 & 0.108462 \\ 
         \makecell[l]{Q5 (80 \%-100 \%)} & 0.750752 & 0.036253 & 0.048288 \\ 
        \bottomrule
    \end{tabular}
    
    \label{tab:building_function_size}
\end{table}

\newpage
\section{Estimation of SVI Covered Population}
\label{sec:sample:appendix_G}
The research aims to understand to what extent the SVI coverage of residential buildings also achieves coverage of the total population within the study area. The study uses the proportion of residential buildings visible in the SVI at the H3 level-9 grid as a rough representation of the SVI coverage rate of the population in each grid. At the same time, Meta 30 m precision population data is used to estimate the number of people in each H3 grid. The proportion of total population covered in SVI can be calculated based on the Formula \ref{eq:simplified_population_covered}. Based on the proportion of residential buildings covered, it is estimated that approximately 66.2 \% of the total population in the study area can be covered by SVI.

\begin{equation}
\textit{Total Population Covered in SVI} = \sum_{n} \left( CoC-A_r \times P_n \right)
\label{eq:simplified_population_covered}
\end{equation}

where:
\begin{itemize}
    \item $CoC-A_r$ represents the completeness of SVI coverage on residential buildings in the $n^{th}$ grid.
    \item $P_n$ denotes the population in the $n^{th}$ grid.
\end{itemize}

\end{document}